\pdfoutput=1

\documentclass[final]{cvpr}

\usepackage{times}
\usepackage{epsfig}
\usepackage{graphicx}
\usepackage{amsmath}
\usepackage{amssymb}

\usepackage{multirow}
\usepackage{comment}
\usepackage{xcolor,colortbl} 
\usepackage{soul}

\usepackage[ruled,vlined,linesnumbered]{algorithm2e}

\usepackage[abbreviations]{foreign}
\usepackage{enumitem}

\usepackage{bm}
\usepackage{dsfont}

\makeatletter
\@namedef{ver@everyshi.sty}{}
\makeatother
\usepackage{tikz}
\usepackage{pgfplots}
\usetikzlibrary{pgfplots.groupplots}
\usepgfplotslibrary{groupplots}

\usepackage{fp}
\usepackage{xfp}

\usepackage{multicol}
\usepackage{booktabs}       
\usepackage{multirow}

\usepackage{array}
\newcolumntype{H}{>{\setbox0=\hbox\bgroup}c<{\egroup}@{}}

\usepackage[pagebackref=true,breaklinks=true,colorlinks,bookmarks=false,citecolor=blue]{hyperref}

\def\cvprPaperID{5385} 
\def\confYear{CVPR 2021}

\newcommand{\ketopic}[1]{\noindent\textbf{#1:}}
\newcommand{\keheading}[1]{\noindent\underline{\textbf{#1}}\\}
\newcommand\kebullet[1][.5]{\mathbin{\vcenter{\hbox{\scalebox{#1}{$\bullet$}}}}}

\newcommand{\ketrim}[1]{\FPeval{\result}{round(#1,2)}\result}
\newcommand*{\rom}[1]{\uppercase\expandafter{\romannumeral #1\relax}}

\begin{document}

\title{Knowledge Evolution in Neural Networks}

\author{Ahmed Taha
	\and		Abhinav Shrivastava	\\ 	University of Maryland, College Park	\and 		Larry Davis
}



\maketitle

\begin{abstract}


Deep learning relies on the availability of a large corpus of data (labeled or unlabeled). Thus, one challenging unsettled question is: how to train a deep network on a relatively small dataset? To tackle this question, we  propose  an evolution-inspired training approach to boost performance on relatively small datasets. The knowledge evolution (KE)  approach splits a deep network into two hypotheses: the fit-hypothesis and the reset-hypothesis. We iteratively evolve the knowledge inside the fit-hypothesis by perturbing the reset-hypothesis for multiple generations. This approach not only boosts performance, but also learns a slim network with a smaller inference cost. KE integrates seamlessly with both vanilla and residual convolutional networks. KE reduces both overfitting and the burden for data collection.





We evaluate KE on various network architectures  and loss functions. We evaluate KE using relatively small datasets  (e.g., CUB-200) and randomly initialized deep networks. KE achieves an absolute 21\% improvement margin on a state-of-the-art baseline. This performance improvement is accompanied by a relative 73\% reduction in inference cost.  KE achieves state-of-the-art results on classification and metric learning benchmarks.  Code available  at \textit{http://bit.ly/3uLgwYb}

%


\end{abstract}

\section{Introduction}
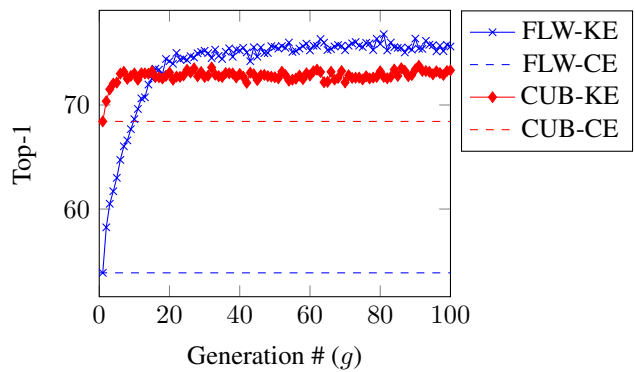
\begin{figure}
	\centering
	\begin{tikzpicture}
		\begin{axis}[
			xlabel=Generation \# ($g$),
			xmin=0,
			xmax=100,
			width=0.75\linewidth,
			legend pos=outer north east,
			y label style={at={(axis description cs:0.05,.5)}},
			ylabel=Top-1]
			\addplot[color=blue,mark=x] coordinates {
				(1, 53.87)(2, 58.24)(3, 60.51)(4, 61.71)(5, 63.00)(6, 64.72)(7, 66.03)(8, 66.60)(9, 67.68)(10, 68.64)(11, 69.62)(12, 70.63)(13, 70.74)(14, 71.96)(15, 72.52)(16, 73.49)(17, 73.40)(18, 72.96)(19, 74.42)(20, 74.21)(21, 73.96)(22, 75.00)(23, 74.37)(24, 74.48)(25, 74.32)(26, 74.69)(27, 74.56)(28, 74.97)(29, 75.10)(30, 75.21)(31, 74.98)(32, 74.51)(33, 75.32)(34, 74.72)(35, 74.42)(36, 75.11)(37, 75.44)(38, 74.63)(39, 75.32)(40, 75.24)(41, 75.50)(42, 75.05)(43, 74.21)(44, 75.55)(45, 74.64)(46, 75.53)(47, 74.98)(48, 75.00)(49, 75.65)(50, 75.57)(51, 75.42)(52, 75.66)(53, 75.49)(54, 76.00)(55, 75.10)(56, 75.18)(57, 75.42)(58, 75.65)(59, 75.89)(60, 75.24)(61, 75.70)(62, 75.68)(63, 76.34)(64, 75.99)(65, 75.28)(66, 75.37)(67, 75.62)(68, 75.76)(69, 75.68)(70, 75.26)(71, 75.70)(72, 75.92)(73, 76.05)(74, 75.70)(75, 75.57)(76, 75.97)(77, 75.92)(78, 75.28)(79, 76.36)(80, 75.87)(81, 76.80)(82, 75.24)(83, 76.00)(84, 75.55)(85, 75.49)(86, 75.49)(87, 75.05)(88, 75.52)(89, 75.32)(90, 76.38)(91, 75.42)(92, 75.39)(93, 76.17)(94, 75.79)(95, 75.74)(96, 75.16)(97, 75.70)(98, 75.18)(99, 75.78)(100, 75.62)
			};
			\addplot[color=blue,	dashed] coordinates {
				(0, 53.87)(100, 53.87)
			};
			\addplot[color=red,mark=diamond*] coordinates {
				(1, 68.42)(2, 70.36)(3, 71.51)(4, 72.06)(5, 72.15)(6, 73.01)(7, 73.19)(8, 72.51)(9, 72.79)(10, 73.03)(11, 72.43)(12, 73.05)(13, 73.03)(14, 73.00)(15, 73.24)(16, 72.69)(17, 72.67)(18, 72.55)(19, 72.79)(20, 73.12)(21, 72.43)(22, 72.72)(23, 72.70)(24, 72.84)(25, 73.38)(26, 73.12)(27, 73.52)(28, 72.76)(29, 73.17)(30, 72.69)(31, 72.82)(32, 73.60)(33, 72.91)(34, 72.77)(35, 72.69)(36, 73.07)(37, 73.05)(38, 73.03)(39, 72.65)(40, 72.98)(41, 72.72)(42, 72.13)(43, 73.10)(44, 72.95)(45, 72.39)(46, 72.86)(47, 72.89)(48, 72.77)(49, 72.72)(50, 72.67)(51, 72.57)(52, 73.19)(53, 72.88)(54, 72.34)(55, 72.43)(56, 72.53)(57, 72.82)(58, 72.48)(59, 73.05)(60, 72.84)(61, 73.24)(62, 73.39)(63, 73.17)(64, 72.19)(65, 72.22)(66, 73.19)(67, 72.44)(68, 72.60)(69, 73.26)(70, 72.17)(71, 72.86)(72, 72.60)(73, 72.62)(74, 72.63)(75, 72.46)(76, 72.76)(77, 72.55)(78, 72.63)(79, 72.69)(80, 73.24)(81, 72.70)(82, 72.57)(83, 72.70)(84, 73.41)(85, 72.79)(86, 73.53)(87, 73.29)(88, 72.77)(89, 72.88)(90, 73.43)(91, 73.77)(92, 73.27)(93, 73.26)(94, 73.19)(95, 72.95)(96, 73.24)(97, 73.07)(98, 72.89)(99, 73.26)(100, 73.33)
			};
			\addplot[color=red,dashed] coordinates {
				(0, 68.42)(100, 68.42)
			};
			
			\legend{FLW-KE,FLW-CE,CUB-KE,CUB-CE}
		\end{axis}
	\end{tikzpicture}
	\caption{Classification performance on Flower-102 (FLW) and CUB-200 (CUB) datasets trained on a randomly initialized ResNet18. The horizontal dashed-lines denote a SOTA cross-entropy (CE) baseline~\cite{yun2020regularizing}. The marked-curves show our approach (KE) performance across generations. The $100^{\text{th}}$ generation KE-$N_{100}$ achieves absolute $21\%$ and $5\%$  improvement margins over the Flower-102 and CUB-200 baselines, respectively.}
	\label{fig:intro_performance}
	\vspace{0.20in}
\end{figure}

\vspace{0.10in}
Gene transfer is the transfer of genetic information from a parent to its offspring. Genes encode genetic instructions (knowledge) from ancestors to descendants. The ancestors do not necessarily have better knowledge; yet, the evolution of knowledge across generations promotes a better learning curve for the descendants. In this paper, we strive to replicate this process for deep networks.  We encapsulate a deep network's knowledge inside a subnetwork, dubbed the fit-hypothesis $H^\triangle$. Then, we pass the fit-hypothesis's knowledge from a parent network to its offspring (next deep network generation). We repeat this process iteratively and demonstrate a significant performance improvement in the descendant networks as shown in \autoref{fig:intro_performance}. 


The lottery ticket literature~\cite{frankle2018lottery,zhou2019deconstructing,morcos2019one,ramanujan2020s,girish2020lottery} regards a dense network as a set of hypotheses  (subnetworks). Zhou~\etal~\cite{zhou2019deconstructing} propose a sampling-based approach, while  Ramanujan~\etal ~\cite{ramanujan2020s} propose an optimization-based approach, to identify the best \textit{randomly-initialized} hypothesis. This hypothesis may be called the lottery ticket, but it is still limited by its random initialization. In this paper, we pick a random hypothesis, with inferior performance, and iteratively evolve its knowledge.

\begin{figure*}[t]
\centering
\scriptsize
\includegraphics[width=0.9\linewidth]{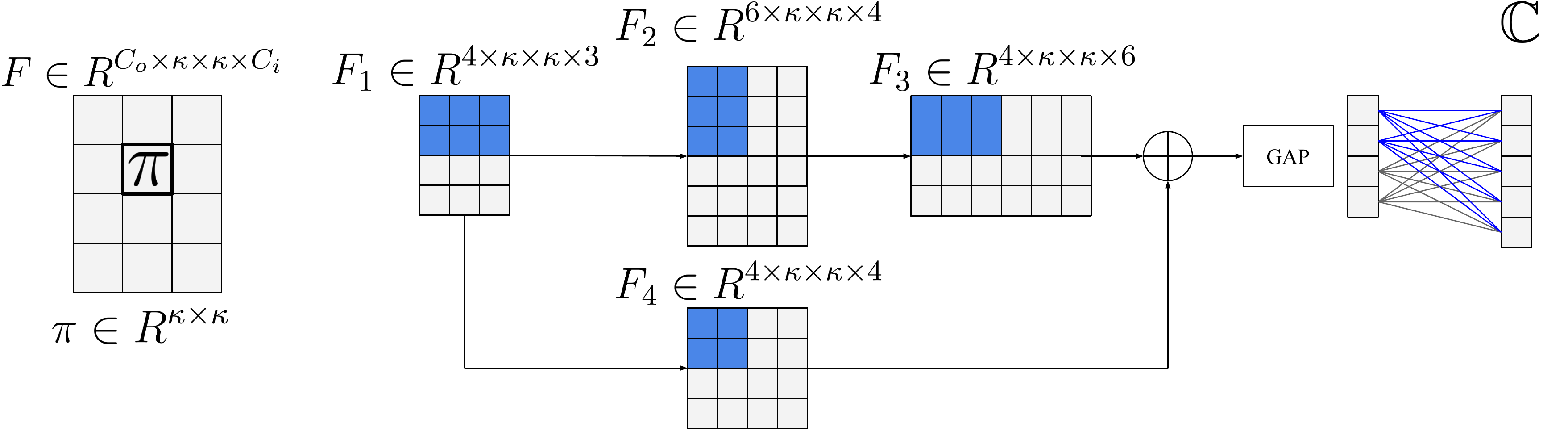}
\caption{A split network illustration using a toy residual network. (Left) A convolutional filter $F$ with $C_i=3$ input, $C_o=4$ output channels, and 2D kernels (\eg $\pi \in R^{3 \times 3}$). (Center-Right) A toy residual network $N$ with a three-channel input (\eg RGB image) and  a five-logit output ($\mathbb{C}=5$). GAP denotes a global average pooling layer while $\bigoplus $ denotes the add operation. We split $N$ into a fit-hypothesis $H^\triangle$ (dark-blue) and a reset-hypothesis $H^\triangledown$ (light-gray). The fit-hypothesis $H^\triangle$ is a slim network that can be extracted from the dense network $N$ to perform inference efficiently. The paper appendix shows the dimensions of a fit-hypothesis in the ResNet18 architecture.}
\label{fig:ke_overview}

\end{figure*}

The \underline{main contribution} of this paper is an evolution-inspired training approach. To evolve knowledge inside a deep network, we split the network into two hypotheses (subnetworks): the fit-hypothesis $H^\triangle$ and the reset hypothesis $H^\triangledown$ as shown in~\autoref{fig:ke_overview}.  We evolve the knowledge inside $H^\triangle$ by re-training the network for multiple generations. For every new generation, we perturb the weights inside $H^\triangledown$ to encourage the $H^\triangle$ to learn an independent representation. This knowledge evolution approach boosts performance on relatively small datasets and promotes a better learning curve for descendant networks. Our intuitions are presented in~\autoref{sec:ke_intuitions} and empirically validated in~\autoref{sec:ablation_study}.

The knowledge evolution (KE) approach requires network-splitting. If we split the weights of a neural network into two hypotheses ($H^\triangle$ and $H^\triangledown$) \textit{randomly}, KE will boost performance. This emphasizes the generality of our approach. Furthermore, we propose  a \underline{\textbf{ke}}rnel-\underline{\textbf{l}}evel convolutional-aware \underline{\textbf{s}}plitting (KELS) technique to reduce inference cost. KELS  is  a splitting technique tailored for convolutional neural networks (CNNs). KELS splits a CNN such that the fit-hypothesis $H^\triangle$ is a slim independent network with a smaller inference cost as shown in~\autoref{fig:ke_overview}. The KELS technique supports both vanilla CNNs (AlexNet and VGG) and modern residual networks.

KE supports various network architectures and loss functions. KE  integrates seamlessly with other regularization techniques (\eg, label smoothing). While KE increases  the training time, the KELS technique reduces the inference cost significantly. Most importantly, KE mitigates overfitting on relatively small datasets, which in turn reduces the burden for data collection. Our community takes natural images for granted because they are available publicly. However, for certain applications, such as autonomous navigation and medical imaging, the data collection process is expensive even when labeling is not required.

In summary, the key contributions of this paper are: 

\begin{enumerate}[noitemsep]
	\item A training approach, knowledge evolution (KE), that boosts the performance of deep networks on relatively small datasets (\autoref{sec:ke_training}). We evaluate KE using both classification (\autoref{sec:exp_cls}) and metric learning (\autoref{sec:exp_ret}) tasks. KE achieves SOTA results.
	\item A network splitting technique, KELS, which learns a slim network automatically while training a deep network (\autoref{sec:split_nets}). KELS supports a large spectrum of CNNs and introduces neither hyperparameters nor regularization terms. Our ablation studies (\autoref{sec:ablation_study}) demonstrate how KELS reduces inference cost significantly.
	
\end{enumerate}

\section{Related Work}\label{sec:related_work}
This section compares knowledge evolution (KE) with two prominent training approaches: Born-Again Networks (BANs)~\cite{furlanello2018born} and  Dense-Sparse-Dense (DSD)~\cite{han2016dsd}. In the paper appendix, we compare KELS with the pruning literature~\cite{lecun1990optimal,hassibi1993second,han2015learning,han2015deep,li2016pruning,wen2016learning,zhou2016less,luo2017thinet,liu2017learning,yu2018nisp,huang2018condensenet}



DSD~\cite{han2016dsd} starts with a dense-phase to learn connections' weights and importance. Then, the sparse-phase prunes the unimportant connections  and resumes training given a sparsity constraint. The final dense-phase removes the sparsity constraint, re-initializes the pruned connections,  and trains the entire dense network. KE differs from DSD in multiple ways: (1) DSD masks (prunes) individual weights, while KE masks complete convolution kernels. Thus, DSD delivers dense networks, while KE delivers both dense and slim networks. (2) KE introduces the idea of a fit-hypothesis to encapsulate a network's knowledge and to evolve this knowledge across generations.





BANs~\cite{furlanello2018born} is a knowledge-distillation based approach. Similar to KE, BANs trains the same architecture iteratively. However, to transfer knowledge between successive networks, BANs uses the class-logits distribution, while KE uses the networks' weights. This explains why BANs uses the teacher-student terminology while KE uses the parent-sibling terminology. This difference is important because (1) training a teacher network, which teaches future students, requires a large corpus of data (labeled or not). In contrast, KE acknowledges the deficiency of a parent network trained on a small dataset; (2) BANs randomly initializes student networks while KE leverages the knowledge of a  parent network to initialize the next generation.


We distance our work from neural architecture search (NAS) literature~\cite{zoph2016neural,liu2018progressive} such as Neural Rejuvenation~\cite{qiao2019neural} and MorphNet~\cite{gordon2018morphnet}. We assume the network's connections and the number of parameters are fixed.


\section{Knowledge Evolution}
In this section, we present (1) the knowledge evolution (KE) approach (\autoref{sec:ke_training}), (2) various network-splitting techniques (\autoref{sec:split_nets}), (3) intuitions behind KE (\autoref{sec:ke_intuitions}), and (4) how we evaluate KE (\autoref{sec:evaluation_tasks}).



\subsection{The Knowledge Evolution Training Approach}\label{sec:ke_training}
We first introduce our notation. We assume a deep network $N$  with $L$ layers. The network $N$ has convolutional filters $F$, batch norm $Z$, and fully connected layers with weight $W$, bias $B$ terms.

The Knowledge evolution (KE) approach starts by \textit{conceptually} splitting  the deep network $N$ into two exclusive hypotheses (subnetworks): the fit-hypothesis $H^\triangle$ and the reset-hypothesis $H^\triangledown$ as shown in~\autoref{fig:ke_overview}. These hypotheses are outlined by a binary mask $M$; 1 for  $H^\triangle$ and 0 for $H^\triangledown$,~\ie $H^\triangle=M N$ and $ H^\triangledown =  (1-M) N$. We present various splitting techniques in \autoref{sec:split_nets}. After outlining the hypotheses, the network $N$ is initialized randomly,~\ie both $H^\triangle$ and $H^\triangledown$ are initialized randomly. We train $N$ for $e$ epochs and refer to the trained network as the first generation $N_1$, where $H_1^\triangle  = M N_1$ and  $ H_1^\triangledown = (1-M)  N_1$.

To learn a better network (the next generation), we  (1) \textbf{re-initialize} the network $N$ using $H_1^\triangle$, then (2) \textbf{re-train} $N$ to learn $N_2$. First, the network $N$ is \textbf{re-initialized} using the convolutional filters $F$ and weights $W$ in the fit-hypothesis $H_1^\triangle$ from $N_1$, while the rest of the network ($H^\triangledown$) is initialized randomly. Formally, we re-initialize each layer  $l$, using Hadamard product, as follows

\begin{equation}\label{eq:transfer_fit_hypothesis}
	F_l = M_l  F_l  + \left(1-M_l\right)   F^r_l,
\end{equation}
where $F_l$ is a convolutional filter at layer $l$,  $M_l$ is the corresponding binary mask and $F^r_l$ is a randomly initialized tensor. These three tensors ($F_l$, $F^r_l$, and $M_l$) have the same size ($\in R^{C_o \times \kappa \times \kappa \times C_i}$). $F^r_l$ is initialized using the default initialization distribution. For example, PyTorch uses Kaiming uniform~\cite{he_delving_2015} for convolution layers.

Similarly, we re-initialize the weight $W_l$ and bias $B_l$ through their corresponding binary masks. Modern architectures have bias terms in the single last fully connected layer only ($B\in R^{\mathbb{C}}$). Thus, for these architectures, all bias terms belong  to the fit-hypothesis,~\ie $B \subset  H^\triangle$. We transfer the learned batch norm $Z$ across generations without randomization.




After re-initialization, we \textbf{re-train} $N$ for $e$ epochs to learn the second generation $N_2$. To learn better networks, we repeatedly \textbf{re-initialize} and \textbf{re-train} $N$ for $g$ generations. Basically, we  transfer  knowledge (convolutional filters and weights) from one generation to the next through the fit-hypothesis $H^\triangle$. It is important to note that (1) the contribution of a network-generation ends immediately-after initializing the next generation,~\ie each generation is trained independently, (2) After training a new generation, the weights inside both hypotheses change,~\ie  $H^\triangle_1\ne H^\triangle_2$ and $H^\triangledown_1\ne H^\triangledown_2$, and (3) all network generations are trained using the exact hyperparameters,~\ie same number of  epochs, optimizer, learning rate scheduler, etc. 


\subsection{Split-Networks}\label{sec:split_nets}
\begin{figure}[t]
	\centering
	\scriptsize
	\includegraphics[width=0.85\linewidth]{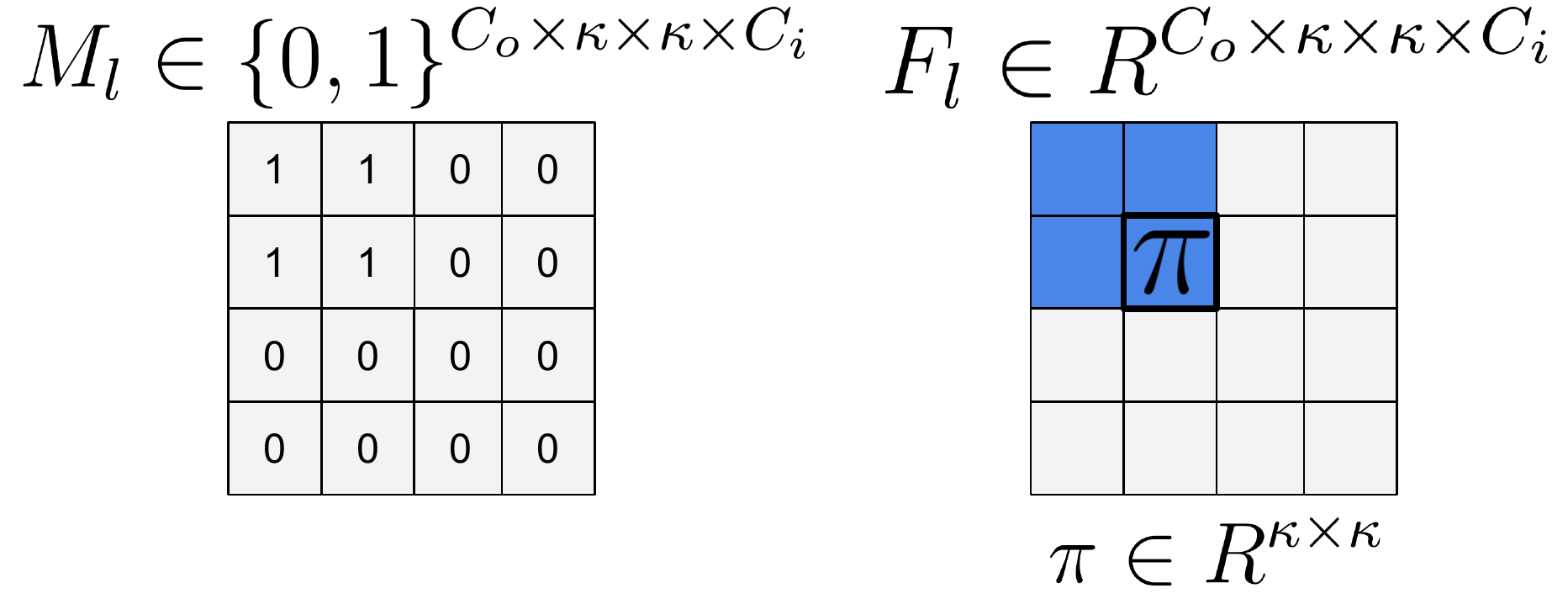}
	\caption{The KELS technique for CNNs. Given a split-rate $s_r$ and a convolutional filter $F_l$ at a layer $l$, the binary split-mask $M_l$ outlines the first  $\left\lceil  s_r \times C_i \right\rceil$ kernels inside the first $\left\lceil  s_r\times C_o \right\rceil  $ filters. In this example, $C_o=C_i=4$ and $s_r=0.5$. Through KELS, the binary mask $M$ outlines the fit-hypothesis $H^\triangle$ such that it is a slim network inside a dense network. The slim network $H^\triangle$ is equivalent to a dense network with $(1-s_r^2)$ sparsity.}
	\label{fig:split_net_mask}
\end{figure}

\begin{algorithm}[t]
	\SetAlgoLined
	\KwResult{Both a dense network $N_g$ and a slim network $H_g^\triangle$ outlined by the split mask $M$}
	\tcc{\scriptsize{Set the split masks $M$ for conv and FC layers once and for all.}} 
	\For{layer $l$ \KwTo $L$}{
		\uIf {\textup{is\_conv($l$)}}
		{
			$C_o$, $\kappa$, \_, $C_i$ = $F_l$.shape\;
			$M_l$ = zeros(($C_o,\kappa,\kappa,C_i$))\;
			\uIf{$C_i$ == 3} {
				$M_l[:C_o\times s_r,:,:,:] = 1$				\tcp*{\scriptsize{First conv}}

			}
			\Else
			{
				$M_l[:C_o\times s_r,:,:,:C_i\times s_r] = 1$\;
			}
		}
		\ElseIf{\textup{is\_fc($l$)}}
		{  
			$C_o$, $C_i$ = $W_l$.shape\tcp*{$\scriptstyle C_o = \mathbb{C}$} 
			$M_l$ = zeros(($C_o,C_i$))\;
			
			$M_l[:,:C_i\times s_r]= 1$\;
		}
	}
	$W,B,Z,F$ are initialized randomly\;
	\For{generation $i$ \KwTo $g$}{
		$N_i\leftarrow$ Train $N$  for $e$ epochs\tcp*{\scriptsize{Learn $\scriptstyle W, B, Z, F$}} 
		\For{layer $l$ \KwTo $L$}{
			\uIf{\textup{is\_conv($l$)}} {
				$F_l^r = \text{rand}(F_l.\text{shape})$\;
				$F_l = M_l  F_l +  (1-M_l)  F_l^r$\;
			}
			\ElseIf{\textup{is\_fc($l$)}} {
				$W_l^r = \text{rand}(W_l.\text{shape})$\;
				$W_l = M_l  W_l +  (1-M_l)  W_l^r$\;
			}
		}
	}
	\caption{The KE training approach splits a dense network $N$, with $L$ layers, into fit and reset hypotheses using a  split-rate $s_r$ and a binary mask $M$. Then, KE trains $N$ for $g$ generations. The network $N$ has convolutional filters $F$, weight $W$, bias $B$,  and batch norm $Z$. We assume a single fully connected layer for simplicity.}
	\label{algo:split_nets}
\end{algorithm}

KE requires network-splitting. We support KE with two splitting techniques: (1) a simple technique to highlight the generality of KE, and (2) an efficient technique for CNNs.


The simple technique is the \textit{\textbf{we}ight-\textbf{l}evel} \textbf{s}plitting (WELS) technique. For every layer $l$, a binary mask $M_l$ splits $l$ into two exclusive parts: the fit-hypothesis $H^\triangle$ and the reset-hypothesis $H^\triangledown$. Given a split-rate $0<s_r<1$, we \textit{randomly} split the weights $W_l \in R^{|W_l|}$ using the mask $M_l\in \{0,1\}^{|W_l|}$, where $|W_l|$ is the number of weights inside layer $l$ and $\text{sum}(M_l)=s_r \times  |W_l|$.  The WELS technique  supports a large spectrum of layers -- fully connected, convolution, recurrent, and graph convolution. This highlights the generality of KE. 

Through WELS, KE boosts the network performance across generations. However, WELS does not benefit from the connectivity of CNNs. Thus, we propose a  splitting technique that not only boosts performance but also reduces inference cost for relatively small datasets. We leverage the CNNs' connectivity and outline the fit-hypothesis $H^\triangle$  such that it is a slim (pruned) network as shown in~\autoref{fig:ke_overview}. Instead of masking individual weights, we mask kernels,~\ie \textit{\textbf{ke}rnel-\textbf{l}evel convolutional-aware} \textbf{s}plitting (KELS) technique. Given a split-rate $s_r$ and a convolutional filter $F_l \in R^{C_o \times \kappa \times \kappa \times C_i}$, KELS outlines the fit-hypothesis to include the first $\left\lceil  s_r \times C_i \right\rceil$ kernels inside the first $\left\lceil  s_r \times C_o \right\rceil$ filters as shown in~\autoref{fig:split_net_mask}. KELS guarantees matching dimensions between consequence convolutional filters. Thus, KELS integrates seamlessly in both vanilla CNNs  (AlexNet and VGG) and modern architectures with residual links. 

For relatively small datasets, the performance of the slim fit-hypothesis $H^\triangle$ reaches the performance of the dense network $N$. In these cases,  $H^\triangle$ not only delivers the dense network's performance but also reduces the inference cost. Through KELS, the slim $H^\triangle$ runs on general purpose hardware,~\ie neither sparse BLAS libraries nor  specialized hardware~\cite{han2016eie} is required. Given a split rate $s_r$, KELS delivers a slim $H^\triangle$ that is equivalent to a dense network $N$ with \textit{approximately} $(1-s_r^2)$ sparsity. It is \textit{approximate} because the network's end-points have $s_r$ sparsity. The first convolutional layer operates on all input channels (\eg RGB) and fully connected layers have $s_r$ sparsity. \autoref{algo:split_nets} summarizes KE  while applying the KELS technique. 


\subsection{Knowledge Evolution Intuitions}\label{sec:ke_intuitions}
\begin{figure}[t]
	\centering
	\scriptsize
	\includegraphics[width=0.6\linewidth]{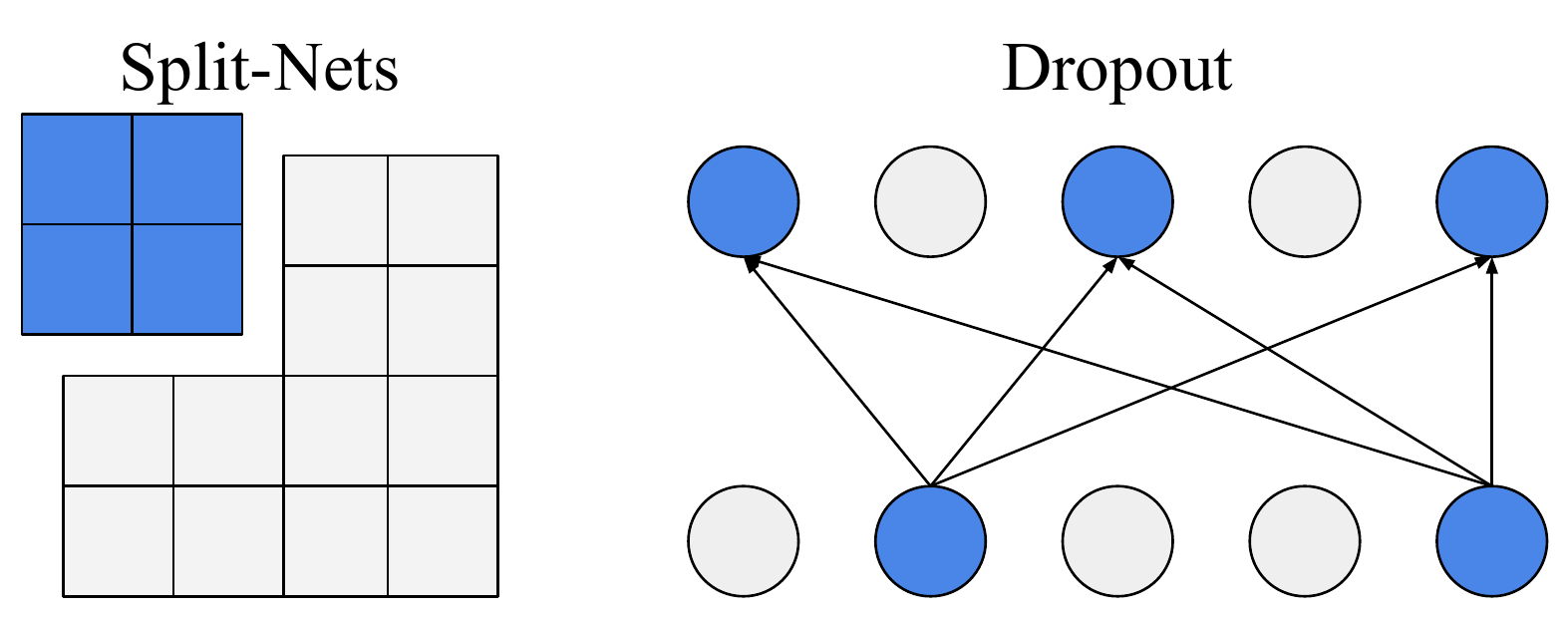}
	\caption{Split-Nets vs Dropout: The reset-hypothesis $H^\triangledown$ and dead neurons are highlighted in gray, while the fit-hypothesis $H^\triangle$ and ``alive'' neurons are highlighted in blue.}
	\label{fig:dropout}
\end{figure}

To understand KE, we give two complementary intuitions. These intuitions do not require the KELS technique. We use KELS for visualization purpose only (\eg\autoref{fig:dropout}). We empirically validate these intuitions in \autoref{sec:ablation_study}.


\begin{figure}[t]
	\centering
	\scriptsize
	\includegraphics[width=0.6\linewidth]{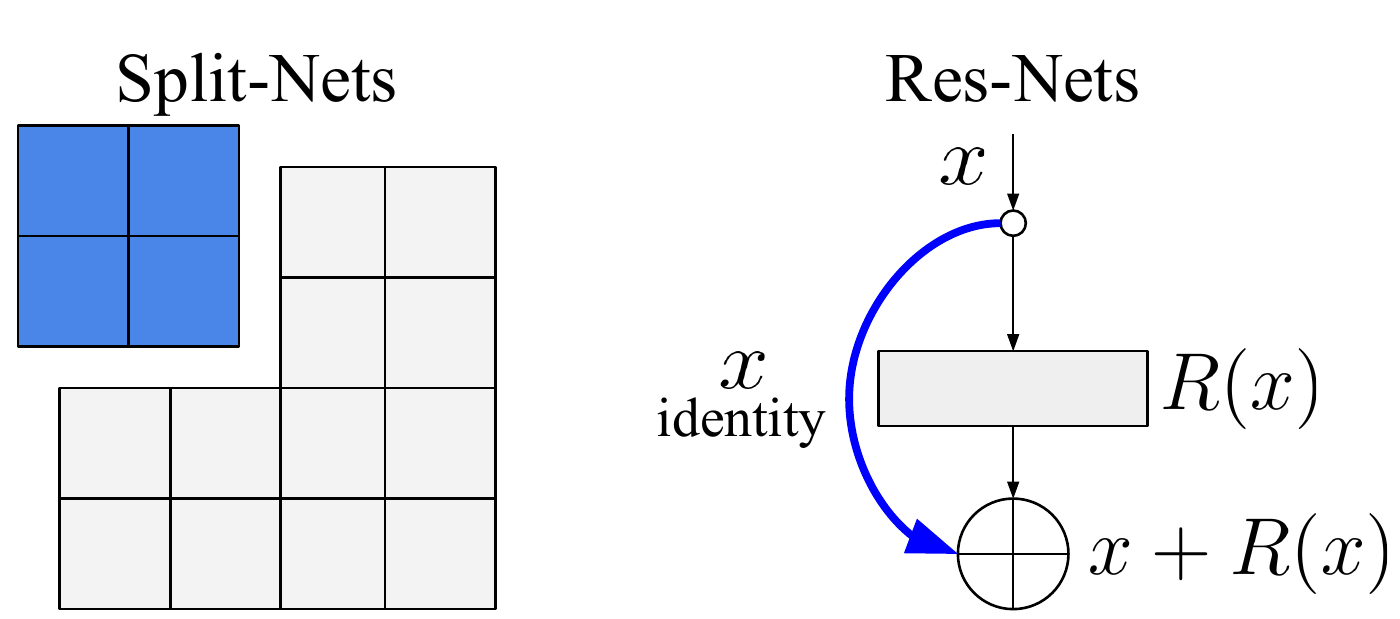}
	\caption{Split-Nets vs Res-Nets: Res-Nets split a network into an identity shortcut (blue) and a residual subnetwork $R(x)$. Split-Nets split a network into a fit-hypothesis $H^\triangle$ (blue) and a reset-hypothesis $H^\triangledown$. By splitting a network into two branches, Res-Net and Split-Net enable a zero-mapping in one of these branches ($R(x)$ and $H^\triangledown$) while keeping the network's depth intact.}
	\label{fig:residual_links}
	%
\end{figure}

\keheading{Intuition \#1: Dropout}
Dropout~\cite{srivastava2014dropout} randomly drops neurons during training as shown in~\autoref{fig:dropout}. This encourages neurons to rely less on each other and to learn independent representations~\cite{cogswell2015reducing}. In KE, we drop the reset-hypothesis $H^\triangledown$ during re-initialization by randomly initializing $H^\triangledown$ before every generation. This encourages  $H^\triangle$ to rely less on  $H^\triangledown$ and to learn an independent representation. We validate this intuition by evaluating the performance of the slim $H^\triangle$ across generations. We observe that the performance of $H^\triangle$ increases as the number of generations increases.

\keheading{Intuition \#2: Residual Network}
Res-Nets set the default mapping, between consecutive layers, to the identity as shown in~\autoref{fig:residual_links}. Yet, from a different perspective, Res-Nets enable a zero-mapping in some subnetworks (residual links) without limiting the network's capacity~\cite{veit2016residual,wu2018blockdrop}. Similarly, KE enables a zero-mapping in the reset-hypothesis $H^\triangledown$ by re-using the fit-hypothesis $H^\triangle$ across generations. After the first generation $N_1$, $H^\triangle$ is always closer to convergence compared to $H^\triangledown$ that contains random values. Thus, KE encourages new generations to evolve the previous-generations' knowledge inside the fit-hypothesis $H^\triangle$ and suppress $H^\triangledown$.



We validate this intuition by measuring the mean absolute value inside both hypotheses. We observe that $H^\triangle$ and $H^\triangledown$ have comparable mean values at the first generation $N_1$. However, as the number of generations increases, the mean absolute value inside $H^\triangle$ increases and $H^\triangledown$ decreases. This supports our claim that KE promotes a zero-mapping inside the reset-hypothesis $H^\triangledown$. 

 Please note that Split-Nets have one degree of freedom that Res-Nets omit. Through the split-rate $s_r$, we control the size of the fit and reset hypotheses ($H^\triangle$ and $H^\triangledown$). If the training data is abundant, a large split-rate is better where a Split-Net reverts into a dense Res-Net. However, for relatively small datasets, a small split-rate $s_r$ significantly reduces the inference cost while improving performance. In the paper appendix, we elaborate more on both intuitions.

\subsection{Evaluation Tasks}\label{sec:evaluation_tasks}
We evaluate KE using two supervised tasks: (1) classification and (2) metric learning. The performance of deep networks on small datasets is studied extensively using the classification task~\cite{szegedy2016rethinking,pereyra2017regularizing,dubey2018maximum,chen2018virtual,muller2019does,xu2019data,zhang2019your,zhang2019adacos,yun2020regularizing}. Thus, the classification task provides a rigorous performance benchmark.  The metric learning evaluation highlights the flexibility of our approach and shows the generality of KE beyond mainstream literature that requires class logits.


\begin{figure}[t]
	\centering
	\includegraphics[width=0.4\linewidth]{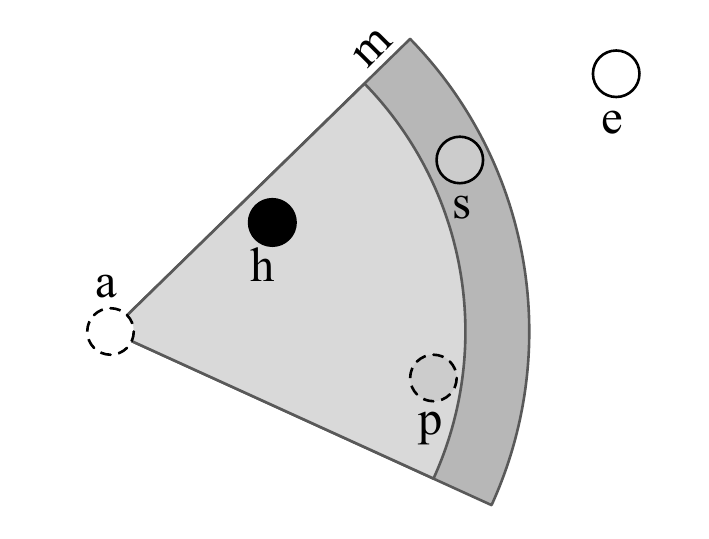}
	\caption{Triplet loss tuple (anchor, positive, negative) and margin $m$. The (h)ard, (s)emi-hard, and (e)asy negatives are highlighted in black, gray, and white, respectively.}
	\label{fig:semi_neg}
\end{figure}

We benchmark KE using both the cross-entropy and the triplet loss. We use these loss functions because most supervised tasks employ one of them. 

\noindent\textbf{Cross-Entropy (CE) Loss:}
We denote $x \in X$ as an input and $y \in Y = \left\{1, ..., \mathbb{C}\right\}$ as its ground-truth label. For a classification network $N$, CE  is defined as follows
\begin{equation}
	\text{CE}_{(x,y)} = -\log \frac{\exp\left(N(x;y)\right)}{\sum^{\mathbb{C}}_{i=1}{ \exp\left(N(x;i)\right) }},
\end{equation}
where $N(x;y)$ denotes the output logit for class $y$ given $x$.

\noindent\textbf{Triplet Loss:} A metric learning network learns an embedding where samples from the same class are close together, while samples from different classes are far apart. To train a metric learning network, we leverage triplet loss for its simplicity and efficiency. Triplet loss is defined as follows
%
\begin{equation}\label{eq:triplet}
\text{TL}_{(a,p,n) \in T} ={ { \left[  { (D_{a,p}-{ D_{a,n} } +m) }  \right]  }_{ + }  },
\end{equation}
where ${ \left[ \kebullet[0.75] \right]  }_{ + }= \max{(0,\kebullet[0.75])}$, $m$ is the margin between classes.  $D_{x_1,x_2}=D(N(x_1),N(x_2))$; $N(\kebullet[0.75]) $ and $D(,)$ are the network's output-embedding and Euclidean distance, respectively. In~\autoref{eq:triplet}, $a$, $p$, and $n$ are the anchor, positive, and negative images in a triplet $(a,p,n)$ from the triplets set $T$.

The performance of triplet loss relies heavily on the sampling strategy. Since we  train randomly initialized networks, we leverage the semi-hard sampling strategy for its stability~\cite{schroff2015facenet,taha2020boosting}. In semi-hard negative sampling, instead of picking the hardest positive-negative samples, all anchor-positive pairs and their corresponding semi-hard negatives are considered. Semi-hard negatives are further away from the anchor than the positive exemplar yet within the banned margin $m$ as shown in~\autoref{fig:semi_neg}. Semi-hard negatives ($n$) satisfy~\autoref{eq:btl_semi_neg}

\begin{equation}\label{eq:btl_semi_neg}
	D_{a,p} < D_{a,n} < D_{a,p} + m.
\end{equation}



\section{Experiments}
In this section, we evaluate KE using classification and metric learning tasks. 

\subsection{Knowledge Evolution on Classification}\label{sec:exp_cls}

\ketopic{Datasets} We evaluate KE using five datasets: Flower-102~\cite{nilsback2008automated}, CUB-200-2011~\cite{wah2011caltech}, FGVC-Aircraft~\cite{maji13fine-grained}, MIT67~\cite{quattoni2009recognizing}, and Stanford-Dogs~\cite{khosla2011novel}. \autoref{tbl:datasets} summarizes the datasets' statistics.

\begin{table}[t]
	\scriptsize
	\centering
	\caption{Statistics of five classification datasets and their corresponding train, validation, and test splits.}
	\begin{tabular}{@{}lccccc@{}}
		\toprule
		& $\mathbb{C}$ & Trn & Val & Tst & Total  \\
		\midrule
				Flower-102~\cite{nilsback2008automated} & 102 & 1020 & 1020 & 6149 &  8189 \\
				CUB-200~\cite{wah2011caltech} & 200 & 5994 & N\slash A & 5794 & 11788 \\
				Aircraft~\cite{maji13fine-grained} & 100 & 3334 & 3333 & 3333 & 10000 \\
				MIT67~\cite{quattoni2009recognizing} & 67 & 5360 & N\slash A & 1340 & 6700 \\
				Stanford-Dogs~\cite{khosla2011novel} & 120 & 12000 & N\slash A & 8580 & 20580 \\
		\bottomrule
	\end{tabular}
	\label{tbl:datasets}
\end{table}


\noindent\textbf{Technical Details:} We evaluate KE using two architectures: ResNet18~\cite{he2016deep,he2016identity} and DenseNet169~\cite{huang2017densely}. These architectures demonstrate the efficiency of KE on modern architectures. All networks are initialized randomly and optimized by stochastic gradient descent (SGD) with momentum 0.9 and weight decay 1e-4. We use cosine learning rate decay~\cite{loshchilov2016sgdr} with an initial learning rate $lr=0.256$. We use batch size $b=32$ and train $N$ for $e=200$ epochs. We use the standard data augmentation technique,~\ie flipping and random cropping. For simplicity, we use the same training settings ($lr,b,e$) for all generations. We report the network accuracy at the last training epoch,~\ie no early stopping.


\ketopic{Baselines} We benchmark KE using the cross-entropy (CE), label-smoothing (Smth) regularizer~\cite{muller2019does,szegedy2016rethinking}, RePr~\cite{prakash2019repr}, CS-KD~\cite{yun2020regularizing}, AdaCos~\cite{zhang2019adacos}, Dense-Sparse-Dense (DSD)~\cite{han2016dsd}, and Born Again Networks (BANs)~\cite{furlanello2018born} introduced in ~\autoref{sec:related_work}:
\begin{itemize}[noitemsep]
	\item \textbf{DSD} determines the duration of each training phase (\# epochs) using the loss-convergence criterion. For small datasets, the loss converges rapidly to zero and some datasets do not have validation splits (see~\autoref{tbl:datasets}). So, we use $e=200$, $e=100$, and $e=100$ epochs for the dense, sparse, dense phases, respectively. We prune each layer  to the default 30\% sparsity.
	
	\item \textbf{AdaCos} maximizes the inter-class angular margin by dynamically scaling the cosine similarities between training samples and their corresponding class center. Thus, AdaCos is  a hyperparameter-free feature embedding regularizer.
	\item \textbf{CS-KD} is a knowledge distillation inspired approach that achieves state-of-the-art performance on small datasets. It distills the logits distribution between different samples from the same class. Thus, it mitigates overconfident predictions and reduces intra-class variations. We set CS-KD's hyperparameters $T=4$ and $\lambda_{\text{cls}}=3$ in all experiments.
	\item \textbf{RePr} is similar to DSD, but instead of pruning weights, RePr prunes \textit{redundant} convolutional filters. Prakash~\etal~\cite{prakash2019repr}  recommend repeating the dense-sparse-dense phases three times. Since we train $N$ for $e=200$ epochs, we set RePr's hyperparameters $S1=50$ and $S2=10$. We use the default sparsity rate (prune rate) $p=30\%$.
\end{itemize}






\begin{table}[t]
	\scriptsize
	\centering
	\caption{Quantitative classification evaluation (Top-1 $\uparrow$) using ResNet18 with KELS. $N_g$ denotes the performance of the $g^{\text{th}}$ network generation. The first generation $N_1$ is both a baseline and a starting point for KE. As the number of generations increases, KE boosts performance.}
			\begin{tabular}{@{}l c c c c c@{}}
			\toprule
			Method & Flower & CUB & Aircraft & MIT & Dog\\
			\midrule
			CE + AdaCos& \bf55.45 & \bf 62.48 & \bf 57.06 & 56.25 & \bf65.34 \\
			CE + RePr& 41.90 & 42.88 & 39.43 & 46.94 & 50.39\\
			CE + DSD &  51.39 & 53.00  & 57.24  & 53.21 &  63.58\\
			CE + BANs-$N_{10}$ &  48.53 &	53.71 &	53.19	& 55.65  & 64.16\\
			CE ($N_1$) & \ketrim{48.48} & \ketrim{53.57} & \ketrim{51.28} & \ketrim{55.28} & 63.83\\
			CE + KE-$N_3$ \textbf{(ours)}&  52.53 & 56.73 & 52.53  & 57.44  & 64.28\\
			CE + KE-$N_{10}$ \textbf{(ours)} & 56.15 & 58.11  & 53.21  & \bf 58.33  & 64.56 \\
			\midrule
			Smth ($N_1$)& 50.97 & 59.75 & 55.00 &  57.74 & 65.95\\
			Smth + KE-$N_3$ \textbf{(ours)} & 56.87 & 62.88 & 57.47 & 58.78 & 66.91\\
			Smth + KE-$N_{10}$ \textbf{(ours)}& \bf62.56 & \bf66.85 & \bf60.03 & \bf60.42 & \bf 67.06\\
			\midrule
			CS-KD ($N_{1}$)& 55.10	& 67.71 &	58.15	& 57.37 & 69.60\\
		CS-KD + KE-$N_{3}$ \textbf{(ours)}& 61.74	& 71.63 & 	\bf 59.97	& \bf 58.41 & 70.62\\
		CS-KD + KE-$N_{10}$ \textbf{(ours)}& \bf69.88	& \bf 73.39 &	59.08	& 57.96 & \bf 70.81\\
			\bottomrule
		\end{tabular}
	\label{tbl:resnet18}
\end{table}

\begin{table}[t]
	\scriptsize
	\centering
		\caption{Quantitative evaluation using DenseNet169 with WELS.}
			\begin{tabular}{@{}l c c c c c@{}}
			\toprule
			Method & Flower  & CUB & Aircraft & MIT & Dog\\
			\midrule
			CE + AdaCos& 49.96	& \bf62.20	& 56.15	& 50.89	& 65.33 \\
			CE + RePr& 39.75 & 47.01& 36.04 & 49.77 & 55.63\\
			CE + DSD &  48.85 & 56.11  & 53.66  & 58.31 & 65.76\\
			CE + BANs-$N_{10}$ & 44.92	& 57.30 &	52.56	& 57.66 & 65.49\\
			
			CE ($N_1$) & 45.85	& 55.16	& 51.73	& 56.62	& 64.82 \\
			CE + KE-$N_3$ \textbf{(ours)}&  52.44	& 57.75	& 56.70 &	\bf59.67	& 67.06\\
			CE + KE-$N_{10}$ \textbf{(ours)}&  \bf 60.15	& 58.01	& \bf 59.73	&  58.71	& \bf67.75 \\
			\midrule
			Smth ($N_{1}$) 											& 46.34	& 59.93	& 57.74	& 57.81	& 65.12\\
			Smth + KE-$N_{3}$ \textbf{(ours)}& 55.46	& \bf 62.53	& 62.86	& \bf 60.27	& \bf68.21\\
			Smth + KE-$N_{10}$ \textbf{(ours)}& \bf 64.18	& 61.34	& \bf 65.86	& 59.75	&  67.46\\
			\midrule
			CS-KD ($N_{1}$) & 46.97	& 67.32	& 58.87	& 56.62 &	69.83 \\
			CS-KD + KE-$N_{3}$ \textbf{(ours)}& 59.36	& 69.77 &	59.91 &	\bf59.00	& \bf71.70\\
			CS-KD + KE-$N_{10}$ \textbf{(ours)}& \bf 65.27	& \bf 70.36	& \bf61.22	& 57.44	& 70.72 \\
			\bottomrule
		\end{tabular}
	\label{tbl:densenet169}
\end{table}

\ketopic{Results} Tables~\ref{tbl:resnet18} and~\ref{tbl:densenet169} present quantitative classification evaluation using ResNet18 and DenseNet169, respectively. For ResNet18, we use a split-rate $s_r=0.8$ and KELS,~\ie $\approx36\%$ sparsity. For DenseNet169, we use $s_r=0.7$ and WELS,~\ie $30\%$ sparsity. We report the performance of the dense network $N$ because all baselines learn dense networks. In~\autoref{sec:ablation_study}, we report the slim fit-hypothesis $H^\triangle$ performance and inference cost. Tables~\ref{tbl:resnet18} and~\ref{tbl:densenet169}  present the performance of the first generation ($N_1$) as a baseline, the third generation ($N_3$) as the short-term benefit, and the tenth-generation ($N_{10}$) as the long-term benefit of KE. 

A deeper network achieves higher accuracy when presented with enough training data. However, if the training data is scarce, a deeper network becomes vulnerable to overfitting. This explains why regularization techniques (\eg AdaCos)  deliver competitive performance on the small ResNet18, but degrade on the large DenseNet169. Interestingly, KE remains resilient on the large DenseNet169 and delivers similar, if not superior, performance. 

We applied KE on top of  (1) the cross-entropy loss, (2) the label smoothing (Smth)  regularizer with its hyperparameter~\cite{muller2019does} $\alpha=0.1$, and (3) the CS-KD regularizer. KE is flexible and boosts performance on each baseline. $N_3$ outperforms $N_1$ on all datasets. After reaching a peak, KE's performance fluctuates. Thus, if  $N_3$ outperforms $N_{10}$ marginally, this indicates that KE reached its peak. In~\autoref{fig:intro_performance}, KE reached its peak on CUB-200 after 20 generations, then KE fluctuates for 80 generations without degrading.



Even though RePr seems similar to KE, the following caveat explains RePr's inferior performance. RePr ranks the \textit{redundant} filters across the entire network,~\ie no per-layer ranking. Prakash~\etal~\cite{prakash2019repr} report pruning more filters from deeper layers when training on large datasets. Yet, RePr prunes many filters from earlier layers when training on small datasets. The earlier layers get a small gradient compared to deeper layers; and with small datasets, the earlier filters remain close to their initialization,~\ie no significant difference between earlier filters. Pruning earlier filters cripples the optimization process and achieves an inferior performance. 

Another important difference between KE and RePr is how filters are re-initialized. KE re-initializes the reset-hypothesis randomly. Thus, KE makes no assumptions about the network architecture. In contrast, RePr is designed specifically for CNNs. RePr re-initializes the pruned filters to be orthogonal to both their values before being dropped and the current value of non-pruned filters. RePr uses the QR decomposition on the weights of the filters from the same layer to find the null-space, that is used to find an orthogonal initialization point. Basically, RePr stores the pruned filters to use them for re-initialization. This makes RePr more complex compared to KE. In the paper appendix, we highlight other differences.

Similar to KE, The BANs training approach trains a network for multiple generations. However, BANs transfers knowledge through the class-logits distribution. For small datasets, a teacher's logits distribution resembles the ground-truth labels (one-hot vector) when the loss converges to zero. Thus, BANs achieves  regular cross-entropy performance even after training for 10 generations.




\subsection{Knowledge Evolution on Metric Learning}\label{sec:exp_ret}
\ketopic{Datasets} We evaluate KE using two standard metric learning datasets: CUB-200-2011~\cite{wah2011caltech}, Stanford Cars196~\cite{krause20133d}. 

\ketopic{Evaluation Metrics} For quantitative evaluation, we use the Recall@K metric and Normalized Mutual Info (NMI) on the test split.

\noindent\textbf{Technical Details:} We use the same hyperparameters ($e$, $lr$ scheduler) and optimizer used in the classification experiments. However, the feature embedding $\in R^{d=128}$ is normalized to the unit circle and we use a batch size $b=125$. Each mini-batch contains $25$ different classes and $5$ samples per class. We use a small learning rate $lr=0.0256$ to avoid large fluctuations in the feature embedding during training.


\begin{table}[t]
	\centering
	\scriptsize
		\setlength\tabcolsep{5.50pt} 
\caption{Quantitative retrieval evaluation using standard metric learning datasets and architectures.}
	\begin{tabular}{@{}l   ccc l@{\hspace{1.0\tabcolsep}} ccc@{}}
		\toprule
		   			 &  \multicolumn{3}{c}{ResNet50} && \multicolumn{3}{c}{GoogLeNet}\\
		   			 \cmidrule{2-4} \cmidrule{6-8}
		Datasets &  NMI & R@1 & R@4  && NMI & R@1 & R@4\\
		\midrule
		CUB ($N_{1}$)  &   0.396	& 13.01	&30.37  &  &   0.396	&10.16&	25.71     \\
		CUB + KE-$N_{3}$  \textbf{(ours)}&     0.424	& 17.22	& 36.14     & &    0.418	&13.94 &	33.78   \\
		CUB + KE-$N_{10}$  \textbf{(ours)}&     \bf0.429	& \bf18.25	& \bf39.40    &&  \bf0.419	& \bf15.34	& \bf34.30     \\
		\midrule
		Cars ($N_{1}$)&     0.374 &	11.63	& 28.66     &&  0.319 &	5.29	& 17.94     \\
		Cars + KE-$N_{3}$ \textbf{(ours)} &     0.514	& 34.28	& 60.25     && 0.476	& 24.98	& 50.06     \\
		Cars + KE-$N_{10}$  \textbf{(ours)}&    \bf0.523	& \bf42.36	& \bf68.11  & &    \bf 0.495	& \bf32.63	& \bf58.84    \\
		\bottomrule
	\end{tabular}

\label{tbl:quan_ret}
\end{table}

\ketopic{Results}~\autoref{tbl:quan_ret} presents a quantitative retrieval evaluation using two standard metric learning architectures: ResNet50~\cite{he2016deep,he2016identity} and GoogLeNet~\cite{szegedy2015going}. We use a split-rate $s_r=0.8$ and KELS with both architectures (See the paper appendix on how KELS handles concatenation operations inside GoogLeNet).  As the number of generations increases, the retrieval performance of the dense network increases. Through this experiment, we highlight how KE supports a large spectrum of network architectures and loss functions. Equipped with WELS, we expect KE to spread beyond CNNs. It is straight forward to tweak WELS and impose a regular sparsity, as in KELS, but for non CNNs.




\section{Ablation Study}\label{sec:ablation_study}
This section presents three ablation studies: We  (1)  validate the dropout and Res-Net intuitions (from \autoref{sec:ke_intuitions}), (2) compare WELS and KELS techniques, (3) present the tradeoffs of the split-rate $s_r$.



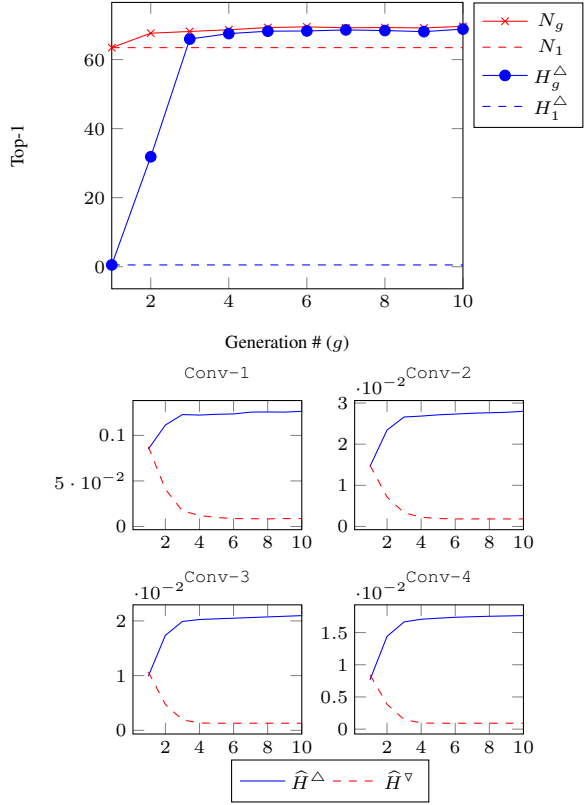
\begin{figure}[t]
	\centering
	\scriptsize
	\begin{tikzpicture}
		\begin{axis}[
			xlabel=Generation \# ($g$),
			xmin=1,
			xmax=10,
			width=0.75\linewidth,
			legend pos=outer north east,
			ylabel=Top-1]
			\addplot[color=red,mark=x] coordinates {
				(1, 63.47)(2, 67.65)(3, 68.15)(4, 68.65)(5, 69.29)(6, 69.48)(7, 69.22)(8, 69.29)(9, 69.16)(10, 69.65)
			};
			\addplot[color=red,	dashed] coordinates {
				(1, 63.47)(15, 63.47)
			};
			\addplot[color=blue,mark=*] coordinates {
				(1, 0.52)(2, 31.85)(3, 65.95)(4, 67.53)(5, 68.23)(6, 68.30)(7, 68.60)(8, 68.42)(9, 68.10)(10, 68.84)
			};
			\addplot[color=blue,	dashed] coordinates {
				(1, 0.52)(15, 0.52)
			};
			\legend{$N_g$,$N_1$,$H^\triangle_g$,$H^\triangle_1$}
		\end{axis}
	\end{tikzpicture}
	
	\begin{tikzpicture}
		\begin{groupplot}[group style = {group size = 2 by 2, horizontal sep = 20pt}, 
			height=3.3cm,
			xmax=10,
			x label style={at={(axis description cs:0.5,-0.05)},anchor=north}
			]
			\nextgroupplot[title=\texttt{Conv-1}, 
			legend style = { legend columns = -1, legend to name = grouplegend,}]
			\addplot[	color=blue,	]
			coordinates {
				(1, 0.08493681252002716)(2, 0.11136557161808014)(3, 0.12285859882831573)(4, 0.12230779230594635)(5, 0.12317689508199692)(6, 0.12356199324131012)(7, 0.12555129826068878)(8, 0.1257077306509018)(9, 0.12555478513240814)(10, 0.126451313495636)
				
			}; 
			\addplot[color=red,dashed]
			coordinates {
				(1, 0.08755726367235184)(2, 0.04093746095895767)(3, 0.017116140574216843)(4, 0.012143289670348167)(5, 0.00989651121199131)(6, 0.008689271286129951)(7, 0.008561183698475361)(8, 0.008298011496663094)(9, 0.008697839453816414)(10, 0.008678192272782326)
				
			};
			\nextgroupplot[title=\texttt{Conv-2}, 
			legend style = { legend columns = -1, legend to name = grouplegend,}]
			\addplot[	color=blue,	]
			coordinates {
				(1, 0.014609038829803467)(2, 0.023442327976226807)(3, 0.026609638705849648)(4, 0.02683161571621895)(5, 0.027137508615851402)(6, 0.02731972560286522)(7, 0.02752663567662239)(8, 0.027643781155347824)(9, 0.02774996869266033)(10, 0.02799667976796627)
				
			}; 
			\addplot[color=red,dashed]
			coordinates {
				(1, 0.014834036119282246)(2, 0.007154650054872036)(3, 0.0033310363069176674)(4, 0.002240509493276477)(5, 0.0019245013827458024)(6, 0.001829108688980341)(7, 0.0018273280002176762)(8, 0.0018276867922395468)(9, 0.0018264935351908207)(10, 0.0018355001229792833)
			};

			\nextgroupplot[title=\texttt{Conv-3}, 
			legend style = { legend columns = -1, legend to name = grouplegend,}]
			\addplot[	color=blue,	]
			coordinates {
				(1, 0.009887644089758396)(2, 0.017364487051963806)(3, 0.019900618121027946)(4, 0.020257702097296715)(5, 0.020382963120937347)(6, 0.020495722070336342)(7, 0.020613806322216988)(8, 0.020733438432216644)(9, 0.02084914781153202)(10, 0.020969342440366745)
				
			}; 
			\addplot[color=red,dashed]
			coordinates {
				(1, 0.010663075372576714)(2, 0.004747464787214994)(3, 0.0019253201317042112)(4, 0.0013440673938021064)(5, 0.0012961566681042314)(6, 0.0012949412921443582)(7, 0.0012960736639797688)(8, 0.0012938895961269736)(9, 0.001295912661589682)(10, 0.0012937574647367)
				
			};
			\nextgroupplot[title=\texttt{Conv-4}, 
			legend style = { legend columns = -1, legend to name = grouplegend,}]
			\addplot[	color=blue,	]
			coordinates {
				(1, 0.007663301192224026)(2, 0.014404826797544956)(3, 0.01665215939283371)(4, 0.01706503890454769)(5, 0.017231592908501625)(6, 0.01737307198345661)(7, 0.01746041141450405)(8, 0.017528627067804337)(9, 0.017581263557076454)(10, 0.017632190138101578)
			}; 		
			\addplot[color=red,dashed]
			coordinates {
				(1, 0.008415102027356625)(2, 0.0038613262586295605)(3, 0.0014980798587203026)(4, 0.0009610414854250848)(5, 0.0009160091285593808)(6, 0.0009169641998596489)(7, 0.0009150828118436038)(8, 0.0009145658113993704)(9, 0.0009150652913376689)(10, 0.0009157087770290673)
			};
			\addlegendentry{$\widehat{H}^\triangle$}		
			\addlegendentry{$\widehat{H}^\triangledown$}	
		\end{groupplot}
		\node[below] at ($(group c1r2.south) +(1.5,-0.25)$) {\pgfplotslegendfromname{grouplegend}}; 
	\end{tikzpicture}
	
	\caption{Quantitative classification evaluation using CUB-200 on VGG11\_bn. The x-axis denotes the number of generations. The fit-hypothesis $H^\triangle$ achieves an inferior performance at $g=1$, but its performance increases as the number of generations increases. $\widehat{H}^\triangle$ and $\widehat{H}^\triangledown$ denote the mean absolute value inside $H^\triangle$ and $H^\triangledown$.}
	\label{fig:quan_cub_vgg11}
\end{figure}

\keheading{(1) Dropout and Res-Net intuitions' validation}
To validate the dropout and Res-Net intuitions, we monitor  the fit and reset hypotheses across generations. According to the \underline{dropout intuition}, the fit-hypothesis should learn an independent representation. The KELS technique enables measuring the fit-hypothesis's performance. In this study, we use the CUB-200 dataset, VGG11\_bn~\cite{simonyan2014very}, and a split-rate $s_r=0.5$.~\autoref{fig:quan_cub_vgg11} (Top) shows the performance of the dense network $N$ and the slim fit-hypothesis $H^\triangle$ for 10 generations. The horizontal dashed lines denote the performance of the first generation ($N_1$ and $H_1^\triangle$). At the first generation, the fit-hypothesis's performance is inferior. Yet, as the number of generations increases, the fit-hypothesis performance increases.~\autoref{tbl:ke_kels_eval} (Top section) presents both the performance and inference cost of both $N$ and $H^\triangle$.

According to the \underline{Res-Net intuition}, the reset-hypothesis should converge to a zero-mapping because, after the first generation ($N_1$), the fit-hypothesis is always closer to convergence.~\autoref{fig:quan_cub_vgg11} shows the mean absolute values ($\widehat{H}^\triangle$ and $\widehat{H}^\triangledown$) inside the fit and reset hypotheses. We present these values inside the first four convolution layers of VGG11\_bn (See paper appendix for all eight conv layers). $\widehat{H}_1^\triangle$ and $\widehat{H}_1^\triangledown$ are comparable at $N_1$. However, as the number of generations increases, $\widehat{H}^\triangle$ increases while $\widehat{H}^\triangledown$ decreases.

\begin{table}[t]
	\centering
	\scriptsize
	\caption{Quantitative evaluation for KELS using the number of both operations (G-Ops) and parameters (millions). $\text{\text{Acc}}_g$ denotes the classification accuracy at the $g^{\text{th}}$ generation. $\blacktriangle_{\text{ops}}$ denotes the relative reduction in the number of operations. $\blacktriangle_{\text{acc}}$ denotes the absolute accuracy improvement on top of the dense baseline $N_1$.}
	\begin{tabular}{@{}lcrccccr@{}}
		\toprule
		& \multicolumn{7}{c}{CUB on VGG11\_bn } \\
		\cmidrule{2-8}
		&$s_r$&   $\text{Acc}_1$ & $\text{Acc}_{10}$  &$\blacktriangle_{\text{acc}}$ & \#Ops & $\blacktriangle_{\text{ops}}$ & \#Param \\
		\midrule
		$N_g $ 												  & \multirow{2}{*}{0.5}&     63.47   &   69.65  &6.1\%&  15.22  & -& 259.16  \\
		$H^\triangle_g $ & &  0.52     &   68.84   &5.3\%&  3.85  &74.7\%&  65.20 \\
		
			\midrule
		& \multicolumn{7}{c}{FLW on ResNet18} \\
		\cmidrule{2-8}
		&$s_r$&   $\text{Acc}_1$ & $\text{Acc}_{100}$  &$\blacktriangle_{\text{acc}}$ & \#Ops & $\blacktriangle_{\text{ops}}$& \#Param  \\
		\midrule
		$N_g$ 										 & \multirow{2}{*}{0.8} &     53.87   &   75.62 & 21.7\%&  3.63  &-& 22.44  \\ 
		$H^\triangle_g $  &&       6.41  &    75.62  & 21.7\% & 2.39&  34.1\% & 14.43  \\
				\midrule
				$N_g$ 									& \multirow{2}{*}{0.5}	  &     52.62   &   74.60  & 21.9\%&  3.63  &-& 22.44  \\
		$H^\triangle_g $  &&     0.37    &   74.60   &21.9\%& 0.96   &  73.5\% & 5.64  \\

		\midrule
			& \multicolumn{7}{c}{CUB on GoogLeNet } \\
		\cmidrule{2-8}
		&$s_r$&   $\text{Acc}_1$ & $\text{Acc}_{10}$  &$\blacktriangle_{\text{acc}}$ & \#Ops &  $\blacktriangle_{\text{ops}}$ & \#Param  \\
		\midrule
		$N_g$ 												& \multirow{2}{*}{0.8}&     64.76  &   72.93  & 8.1\%& 3.00 && 11.59  \\ 
		$H^\triangle_g$  &&   0.64    &    71.67 &6.9\%& 1.98      &34.0\% & 7.54  \\ 		
				\midrule
		$N_g$ 												& \multirow{2}{*}{0.5}&     65.18  &   72.44  & 7.2\%& 3.00 && 11.59  \\
		$H^\triangle_g$  &&      0.50  &   57.23  & -7.9\%& 0.81  & 73.0\% &   3.00 \\
		\bottomrule
	\end{tabular}
\label{tbl:ke_kels_eval}
\vspace{-0.1in}
\end{table}

\keheading{(2) WELS vs KELS techniques}
%
KE requires a network-splitting technique. WELS delivers a dense network $N$ only. Thus, we compare WELS and KELS using $N$.~\autoref{fig:abl_flw_resnet_kels_wels_cs_kd} (Left) compares WELS and KELS using Flower-102, cross-entropy with the CS-KD regularizer~\cite{yun2020regularizing}, ResNet18, and two split-rates ($s_r=\{0.5,0.8\}$). KELS and WELS achieve comparable performance. This is promising because WELS can be applied to any neural network.~\autoref{fig:abl_flw_resnet_kels_wels_cs_kd} (Right) re-assures that KELS delivers high performance while reducing inference cost as shown in~\autoref{tbl:ke_kels_eval} (middle section). The performance of  $H^\triangle_{100}$ matches $N_{100}$ because $H^\triangle$ has enough capacity for the small Flower-102. With $s_r=0.5$, KELS achieves an absolute 21\% improvement margin while reducing inference cost by 73\%.

\keheading{(3) The split-rate $s_r$ tradeoffs}
The split-rate $s_r$ controls the size of the fit-hypothesis; a small $s_r$ reduces  the inference cost. Yet, a small $s_r$ reduces the capacity of $H^\triangle$.~\autoref{fig:abl_cub_googlenet_k} (Left) compares two split-rate ($s_r=\{0.5,0.8\}$) using CUB-200 and GoogLeNet for 10 generations. Both split-rates achieve significant improvement margins on the dense network $N$. However, \autoref{fig:abl_cub_googlenet_k} (Right) shows that the large split-rate $s_r=0.8$ helps the fit-hypothesis $H^\triangle$ to converge faster and to achieve better performance.~\autoref{tbl:ke_kels_eval} (third section) highlights this performance and inference-cost tradeoff. For a large dataset, a large split-rate is required to deliver a slim fit-hypothesis $H^\triangle$ with competitive performance.

\subsection{Discussion}
ImageNet~\cite{deng2009imagenet} will eventually become a toy dataset given the increasing size of deep networks~\cite{coates2013deep,sun2017revisiting,mahajan2018exploring,brown2020language} (\eg GPT-3). To train these large networks, unsupervised~\cite{iscen2018mining,caron2018deep} and self-supervised~\cite{wang2015unsupervised,oord2018representation,he2019momentum,tian2019contrastive} learning mitigate the burden of data annotation. However, these learning approaches still require storing and maintaining a large corpus of data. This is (1) expensive even if neither labeling nor curating is required, (2) impractical for  applications with privacy concerns like medical imaging. KE tackles the problem of training deep networks on relatively small datasets.  KE's main limitation is the training time. It takes $\approx 8$ hours to train 100 generations, $200$ epochs each, on Flower-102 using GTX1080Ti GPU.  This long training time can be reduced by monitoring the performance on a validation split.

\newcommand{\lastpagechartsize}{0.50}

\begin{figure}[t]
	\centering
	\scriptsize
	\begin{tikzpicture}
		\begin{axis}[
			xmin=1,
			xmax=100,
			width=\lastpagechartsize\linewidth,
			y label style={at={(axis description cs:0.15,.5)}},
			legend style={
				at={(0.5,-0.15)},
				anchor=north,
				legend columns=2},
			ylabel=Top-1]
			\addplot[color=red,dashed] coordinates {
				(1, 52.70)(2, 58.01)(3, 60.46)(4, 62.08)(5, 63.21)(6, 63.33)(7, 64.17)(8, 66.32)(9, 66.89)(10, 66.47)(11, 67.75)(12, 68.02)(13, 68.72)(14, 69.56)(15, 70.81)(16, 70.81)(17, 71.78)(18, 72.30)(19, 72.31)(20, 72.99)(21, 73.02)(22, 73.61)(23, 73.51)(24, 73.82)(25, 73.95)(26, 74.37)(27, 74.56)(28, 73.92)(29, 73.46)(30, 74.22)(31, 74.24)(32, 74.00)(33, 74.58)(34, 73.96)(35, 73.90)(36, 75.13)(37, 74.94)(38, 74.09)(39, 74.47)(40, 74.47)(41, 74.26)(42, 74.89)(43, 74.87)(44, 74.82)(45, 74.68)(46, 74.90)(47, 75.19)(48, 75.24)(49, 75.28)(50, 74.90)(51, 74.79)(52, 74.85)(53, 74.51)(54, 76.00)(55, 75.28)(56, 75.32)(57, 75.89)(58, 75.15)(59, 75.49)(60, 74.92)(61, 75.42)(62, 74.89)(63, 75.03)(64, 75.05)(65, 75.39)(66, 75.31)(67, 75.40)(68, 75.44)(69, 75.34)(70, 76.02)(71, 75.50)(72, 76.15)(73, 75.02)(74, 75.60)(75, 75.96)(76, 75.03)(77, 75.50)(78, 74.79)(79, 75.29)(80, 75.86)(81, 75.26)(82, 75.50)(83, 75.97)(84, 75.39)(85, 75.92)(86, 75.47)(87, 75.55)(88, 74.89)(89, 75.83)(90, 76.05)(91, 75.36)(92, 76.18)(93, 75.36)(94, 75.28)(95, 76.42)(96, 76.30)(97, 76.23)(98, 75.68)(99, 75.57)(100, 76.13)
			};\addlegendentry{WELS-$s_r=0.5$}
			\addplot[color=red] coordinates {
				(1, 53.61)(2, 58.61)(3, 60.78)(4, 62.37)(5, 64.10)(6, 65.64)(7, 66.16)(8, 66.97)(9, 68.31)(10, 69.56)(11, 70.00)(12, 70.89)(13, 71.42)(14, 72.41)(15, 72.81)(16, 72.56)(17, 73.28)(18, 73.82)(19, 73.95)(20, 73.70)(21, 75.03)(22, 75.13)(23, 74.45)(24, 73.83)(25, 74.29)(26, 74.66)(27, 75.05)(28, 74.37)(29, 74.29)(30, 74.69)(31, 74.55)(32, 75.16)(33, 74.45)(34, 74.58)(35, 74.61)(36, 74.95)(37, 74.38)(38, 74.34)(39, 75.28)(40, 74.87)(41, 74.56)(42, 75.05)(43, 74.26)(44, 75.34)(45, 74.35)(46, 74.55)(47, 74.08)(48, 74.03)(49, 74.53)(50, 74.69)(51, 74.94)(52, 74.94)(53, 74.72)(54, 75.02)(55, 74.50)(56, 74.82)(57, 74.58)(58, 74.56)(59, 74.68)(60, 74.68)(61, 73.48)(62, 74.64)(63, 74.09)(64, 74.85)(65, 74.77)(66, 74.68)(67, 74.58)(68, 74.92)(69, 74.27)(70, 74.29)(71, 74.72)(72, 73.90)(73, 74.11)(74, 73.82)(75, 74.08)(76, 73.66)(77, 74.82)(78, 73.77)(79, 74.48)(80, 73.66)(81, 74.69)(82, 74.71)(83, 74.43)(84, 74.19)(85, 74.34)(86, 73.80)(87, 74.58)(88, 73.77)(89, 74.69)(90, 73.95)(91, 74.29)(92, 73.72)(93, 74.34)(94, 73.93)(95, 74.42)(96, 73.98)(97, 74.16)(98, 74.17)(99, 74.19)(100, 74.72)
			};\addlegendentry{$s_r=0.8$}
			
			\addplot[color=blue,dashed] coordinates {
				(1, 52.62)(2, 59.60)(3, 61.11)(4, 61.11)(5, 62.22)(6, 63.37)(7, 63.97)(8, 64.73)(9, 65.27)(10, 66.45)(11, 67.00)(12, 67.08)(13, 68.69)(14, 68.69)(15, 69.32)(16, 69.72)(17, 70.08)(18, 70.29)(19, 70.63)(20, 71.41)(21, 71.62)(22, 71.83)(23, 71.97)(24, 72.23)(25, 72.90)(26, 72.88)(27, 72.96)(28, 73.01)(29, 73.64)(30, 73.40)(31, 73.92)(32, 74.14)(33, 73.70)(34, 74.19)(35, 73.95)(36, 74.22)(37, 73.88)(38, 74.61)(39, 74.04)(40, 74.11)(41, 74.61)(42, 74.66)(43, 74.42)(44, 74.09)(45, 74.06)(46, 73.95)(47, 74.69)(48, 73.90)(49, 74.79)(50, 73.75)(51, 73.64)(52, 74.03)(53, 74.68)(54, 74.29)(55, 73.80)(56, 74.40)(57, 74.32)(58, 73.75)(59, 74.16)(60, 74.43)(61, 73.74)(62, 74.19)(63, 74.45)(64, 73.77)(65, 74.11)(66, 73.12)(67, 74.29)(68, 73.58)(69, 73.48)(70, 74.00)(71, 74.21)(72, 73.69)(73, 74.01)(74, 74.53)(75, 74.09)(76, 73.51)(77, 74.64)(78, 74.42)(79, 74.06)(80, 74.34)(81, 74.42)(82, 73.96)(83, 73.90)(84, 74.35)(85, 73.75)(86, 74.61)(87, 74.27)(88, 74.81)(89, 74.50)(90, 74.40)(91, 74.56)(92, 74.74)(93, 74.17)(94, 73.46)(95, 74.32)(96, 74.11)(97, 74.90)(98, 74.04)(99, 75.05)(100, 74.60)
			};\addlegendentry{KELS-$s_r=0.5$}
			
			\addplot[color=blue] coordinates {
				(1, 53.87)(2, 58.24)(3, 60.51)(4, 61.71)(5, 63.00)(6, 64.72)(7, 66.03)(8, 66.60)(9, 67.68)(10, 68.64)(11, 69.62)(12, 70.63)(13, 70.74)(14, 71.96)(15, 72.52)(16, 73.49)(17, 73.40)(18, 72.96)(19, 74.42)(20, 74.21)(21, 73.96)(22, 75.00)(23, 74.37)(24, 74.48)(25, 74.32)(26, 74.69)(27, 74.56)(28, 74.97)(29, 75.10)(30, 75.21)(31, 74.98)(32, 74.51)(33, 75.32)(34, 74.72)(35, 74.42)(36, 75.11)(37, 75.44)(38, 74.63)(39, 75.32)(40, 75.24)(41, 75.50)(42, 75.05)(43, 74.21)(44, 75.55)(45, 74.64)(46, 75.53)(47, 74.98)(48, 75.00)(49, 75.65)(50, 75.57)(51, 75.42)(52, 75.66)(53, 75.49)(54, 76.00)(55, 75.10)(56, 75.18)(57, 75.42)(58, 75.65)(59, 75.89)(60, 75.24)(61, 75.70)(62, 75.68)(63, 76.34)(64, 75.99)(65, 75.28)(66, 75.37)(67, 75.62)(68, 75.76)(69, 75.68)(70, 75.26)(71, 75.70)(72, 75.92)(73, 76.05)(74, 75.70)(75, 75.57)(76, 75.97)(77, 75.92)(78, 75.28)(79, 76.36)(80, 75.87)(81, 76.80)(82, 75.24)(83, 76.00)(84, 75.55)(85, 75.49)(86, 75.49)(87, 75.05)(88, 75.52)(89, 75.32)(90, 76.38)(91, 75.42)(92, 75.39)(93, 76.17)(94, 75.79)(95, 75.74)(96, 75.16)(97, 75.70)(98, 75.18)(99, 75.78)(100, 75.62)
			};\addlegendentry{$s_r=0.8$}
			
		\end{axis}
	\end{tikzpicture}\begin{tikzpicture}
		\begin{axis}[
			xmin=1,
			xmax=100,
			width=\lastpagechartsize\linewidth,
			y label style={at={(axis description cs:0.15,.5)}},
			legend style={
				at={(0.5,-0.15)},
				anchor=north,
				legend columns=1},
			]
			\addplot[color=blue,dashed] coordinates {
				(1, 0.37)(2, 6.57)(3, 53.81)(4, 60.88)(5, 62.21)(6, 63.37)(7, 63.96)(8, 64.77)(9, 65.27)(10, 66.45)(11, 67.00)(12, 67.08)(13, 68.69)(14, 68.69)(15, 69.32)(16, 69.72)(17, 70.08)(18, 70.29)(19, 70.63)(20, 71.41)(21, 71.62)(22, 71.83)(23, 71.97)(24, 72.25)(25, 72.90)(26, 72.88)(27, 72.96)(28, 73.01)(29, 73.64)(30, 73.40)(31, 73.92)(32, 74.14)(33, 73.70)(34, 74.19)(35, 73.95)(36, 74.22)(37, 73.88)(38, 74.61)(39, 74.04)(40, 74.11)(41, 74.61)(42, 74.66)(43, 74.42)(44, 74.09)(45, 74.06)(46, 73.95)(47, 74.69)(48, 73.90)(49, 74.79)(50, 73.75)(51, 73.64)(52, 74.01)(53, 74.68)(54, 74.29)(55, 73.80)(56, 74.40)(57, 74.32)(58, 73.75)(59, 74.16)(60, 74.43)(61, 73.74)(62, 74.19)(63, 74.45)(64, 73.77)(65, 74.11)(66, 73.12)(67, 74.29)(68, 73.58)(69, 73.48)(70, 74.00)(71, 74.21)(72, 73.69)(73, 74.01)(74, 74.53)(75, 74.09)(76, 73.51)(77, 74.64)(78, 74.42)(79, 74.06)(80, 74.34)(81, 74.42)(82, 73.96)(83, 73.90)(84, 74.35)(85, 73.75)(86, 74.61)(87, 74.27)(88, 74.81)(89, 74.50)(90, 74.40)(91, 74.56)(92, 74.74)(93, 74.17)(94, 73.46)(95, 74.32)(96, 74.11)(97, 74.90)(98, 74.04)(99, 75.05)(100, 74.60)
			};\addlegendentry{KELS-$s_r=0.5$}
			\addplot[color=blue,solid] coordinates {
				(1, 6.41)(2, 53.38)(3, 60.51)(4, 61.59)(5, 63.02)(6, 64.73)(7, 66.05)(8, 66.61)(9, 67.68)(10, 68.64)(11, 69.62)(12, 70.63)(13, 70.74)(14, 71.96)(15, 72.52)(16, 73.49)(17, 73.40)(18, 72.96)(19, 74.42)(20, 74.21)(21, 73.96)(22, 75.00)(23, 74.37)(24, 74.48)(25, 74.32)(26, 74.69)(27, 74.56)(28, 74.97)(29, 75.10)(30, 75.21)(31, 74.98)(32, 74.51)(33, 75.32)(34, 74.72)(35, 74.42)(36, 75.11)(37, 75.44)(38, 74.63)(39, 75.32)(40, 75.24)(41, 75.50)(42, 75.05)(43, 74.21)(44, 75.55)(45, 74.64)(46, 75.53)(47, 74.98)(48, 75.00)(49, 75.65)(50, 75.57)(51, 75.42)(52, 75.66)(53, 75.49)(54, 76.00)(55, 75.10)(56, 75.18)(57, 75.42)(58, 75.65)(59, 75.89)(60, 75.24)(61, 75.70)(62, 75.68)(63, 76.34)(64, 75.99)(65, 75.28)(66, 75.37)(67, 75.62)(68, 75.76)(69, 75.68)(70, 75.26)(71, 75.70)(72, 75.92)(73, 76.05)(74, 75.70)(75, 75.57)(76, 75.97)(77, 75.92)(78, 75.28)(79, 76.36)(80, 75.87)(81, 76.80)(82, 75.24)(83, 76.00)(84, 75.55)(85, 75.49)(86, 75.49)(87, 75.05)(88, 75.52)(89, 75.32)(90, 76.38)(91, 75.42)(92, 75.39)(93, 76.17)(94, 75.79)(95, 75.74)(96, 75.16)(97, 75.70)(98, 75.18)(99, 75.78)(100, 75.62)
			};\addlegendentry{KELS-$s_r=0.8$}
			
		\end{axis}
	\end{tikzpicture}
	\caption{Quantitative evaluation for both KELS and WELS using Flower-102 on ResNet18 for 100 generations. The x and y axes denote the number of generations and the top-1 accuracy, respectively. (Left) The classification performance of the dense network $N$. (Right) The performance of the slim fit-hypothesis $H^\triangle$.}
	\label{fig:abl_flw_resnet_kels_wels_cs_kd}
\end{figure}
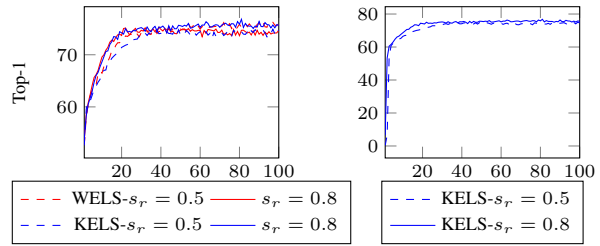

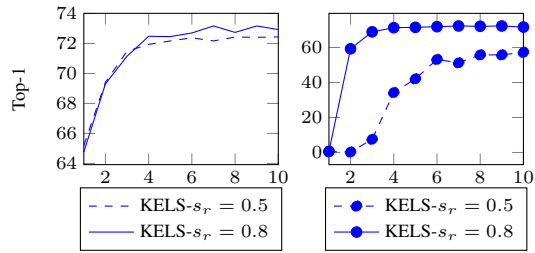
\begin{figure}[t]
	\centering
	\scriptsize
	\begin{tikzpicture}
		\begin{axis}[
			xmin=1,
			xmax=10,
			width=\lastpagechartsize\linewidth,
			y label style={at={(axis description cs:0.15,.5)}},
			legend style={
				at={(0.5,-0.15)},
				anchor=north,
				legend columns=1},
			ylabel=Top-1]
			
			\addplot[color=blue,dashed] coordinates {
				(1, 65.18)(2, 69.41)(3, 71.48)(4, 71.94)(5, 72.17)(6, 72.38)(7, 72.17)(8, 72.43)(9, 72.41)(10, 72.44)
			};\addlegendentry{KELS-$s_r=0.5$}
			
			\addplot[color=blue] coordinates {
				(1, 64.76)(2, 69.30)(3, 71.15)(4, 72.48)(5, 72.46)(6, 72.70)(7, 73.17)(8, 72.74)(9, 73.17)(10, 72.93)
			};\addlegendentry{KELS-$s_r=0.8$}
			
		\end{axis}
	\end{tikzpicture}\begin{tikzpicture}
		\begin{axis}[
			xmin=1,
			xmax=10,
			width=\lastpagechartsize\linewidth,
			y label style={at={(axis description cs:0.15,.5)}},
			legend style={
				at={(0.5,-0.15)},
				anchor=north,
				legend columns=1},
			]
			\addplot[color=blue,dashed,mark=*] coordinates {
				(1, 0.50)(2, 0.26)(3, 7.54)(4, 34.25)(5, 42.07)(6, 53.18)(7, 51.25)(8, 55.77)(9, 55.79)(10, 57.23)
			};\addlegendentry{KELS-$s_r=0.5$}
			\addplot[color=blue,mark=*] coordinates {
				(1, 0.64)(2, 59.27)(3, 68.94)(4, 71.31)(5, 71.53)(6, 71.89)(7, 72.30)(8, 72.13)(9, 72.29)(10, 71.67)
			};\addlegendentry{KELS-$s_r=0.8$}
			
		\end{axis}
	\end{tikzpicture}
	\caption{Quantitative evaluation for different split-rates using CUB-200 on GoogLeNet for 10 generations. 
		(Left) The classification performance of the dense network $N$. (Right) The performance of the slim fit-hypothesis $H^\triangle$.
	}
	
	\vspace{-0.1in}
	\label{fig:abl_cub_googlenet_k}
\end{figure}

\section{Conclusion}
We have proposed knowledge evolution (KE) to train deep networks on relatively small datasets. KE picks a random subnetwork (fit-hypothesis), with inferior performance, and evolves its knowledge. We have equipped KE with a kernel-level convolution-aware splitting (KELS) technique to learn a slim network automatically while training a dense network. Through KELS, KE  reduces the inference cost while boosting performance.   Through the weight-level splitting (WELS) technique, KE supports a large spectrum of architectures. We evaluated KE using classification and metric learning tasks. KE achieves SOTA results. 



\noindent\textbf{Acknowledgments:} This work was partially funded by independent grants from Facebook AI and DARPA SAIL-ON program (W911NF2020009).

{\small
	\bibliographystyle{ieee_fullname}
	\bibliography{split_nets}
}

\clearpage
\section{Appendix}

\newcommand{\beginsupplement}{%
	\setcounter{table}{0}
	\renewcommand{\thetable}{A\arabic{table}}%
	\setcounter{figure}{0}
	\renewcommand{\thefigure}{A\arabic{figure}}%
	\setcounter{section}{0}
	\renewcommand{\thesection}{A\arabic{section}}%
	\setcounter{equation}{0}
	\renewcommand{\theequation}{A\arabic{equation}}%
}


The following appendix-sections extend their corresponding sections in the paper manuscript. For instance, the appendix related-work~\ref{sec:app_related} extends the related-work section in the paper manuscript.

\appendix



\section{Appendix: Related Work}\label{sec:app_related}
The proposed kernel-level convolution-aware splitting (KELS) technique enables the knowledge evolution (KE) approach to learn a slim network with a small inference cost. This signals KE+KELS as a pruning approach. In this section, we compare KE+KELS with the pruning literature. We categorize the pruning  approaches by their pruning-granularity: weights \vs channels \vs  filters. \\




\ketopic{Weight-pruning~\cite{lecun1990optimal,hassibi1993second,han2015learning,han2015deep}} These approaches prune network weights with small absolute magnitude (less salient~\cite{lecun1990optimal}). Weight-pruning reduces the network size, which in turn reduces both DRAM access and energy consumption on mobile devices~\cite{han2015learning}. However, weight-pruning does not reduce the computational costs due to the irregular sparsity after pruning. Accordingly, a weight-pruned network requires sparse BLAS libraries or specialized hardware~\cite{han2016eie}. WELS can be regarded as a weight-pruning technique. However, WELS can be tweaked to reduce both the network size and the computational cost. For instance, we tweaked WELS to propose KELS for CNNs. For a fully connected network (FCN), WELS can split the weights into two independent halves with regular sparsity.  With a regular sparsity, KE delivers a slim, not sparse, FCN.

\ketopic{Channel-pruning~\cite{liu2017learning,yu2018nisp,huang2018condensenet}} Given the limitation of weight-pruning and the complexity of filter-pruning, channel-pruning provides a nice tradeoff between flexibility and ease of implementation. Yet, channel pruning approaches make assumptions. For instance, Liu~\etal~\cite{liu2017learning} require a scaling layer or a batch norm layer; Huang~\etal~\cite{huang2018condensenet} require group convolution support~\cite{krizhevsky2012imagenet}. Accordingly, these~\cite{liu2017learning,huang2018condensenet} are CNN-specific approaches. Furthermore, some channel-pruning approaches (\eg \cite{yu2018nisp}) are applied after training a network. Thus, they do not introduce any performance improvements.



\ketopic{Filter-pruning~\cite{li2016pruning,zhou2016less,luo2017thinet}} KELS belongs to the filter-pruning category.  It is easy to identify unimportant filters, Li~\etal~\cite{li2016pruning} quantify filters' importance using L1-Norm. By removing -- or splitting -- unimportant filters, filter-pruning reduces both the computational cost and the number of parameters. Thus, a filter-pruned network needs neither sparse BLAS libraries nor specialized hardware.  These advantages make filter-pruning appealing. Unfortunately, it is challenging to remove the unimportant filters while maintaining valid network connectivity. For instance, Li~\etal~\cite{li2016pruning} apply filter-pruning on vanilla CNNs (\eg VGG), but require projection-shortcuts to support Res-Nets, and require further modification to support concatenation operations (\eg GoogLeNet). Similarly, ThiNet~\cite{luo2017thinet} suffers on Res-Nets and does not prune the last convolutional layer in all residual blocks. In contrast, KELS supports both vanilla and residual CNNs without bells and whistles.

KE+KELS removes -- or splits -- entire filters. This saves both the number of operations (FLOPs) and parameters (memory). KELS imposes no constraints on the CNN architecture or the loss function. These are key advantages, but KE+KELS has limitations. For instance, KE re-trains a neural network for a large number of generations. This large training cost is not a hurdle for our paper because we tackle the following question: how to train a deep network on a relatively small dataset? 




\section{Appendix: Approach}

\newcommand{\concatsize}{0.80}
\begin{figure}[t]
	\centering
	\scriptsize
	\includegraphics[width=\concatsize\linewidth]{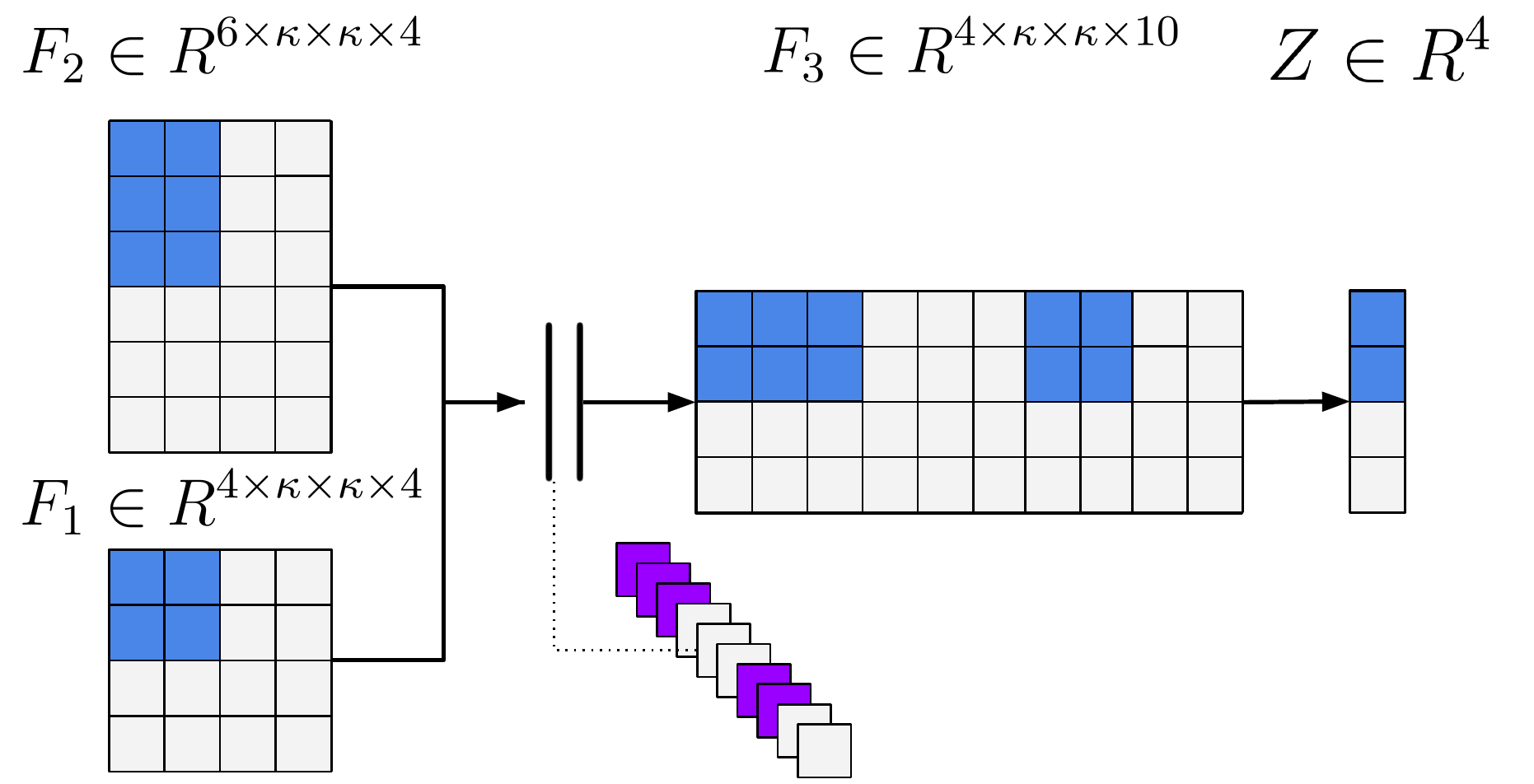}
	\caption{A Split-Net illustration on a toy feature concatenation operation using $s_r=0.5$. $\parallel$ denotes a feature concatenation operation. The dotted line shows the dimension of the  feature map after concatenation~\ie it is not part of the network. In this example, feature concatenation is followed by a convolutional layer $F_3$ then a batch norm layer. This order of operations is employed in \textit{both} GoogLeNet and DenseNet. To split $F_3$  properly (initialize its split-mask $M_3$), we keep references to the preceding convolutional filters ($F_1$ and $F_2$). Through these references, we determine for $F_3$ whether an input channel belongs to the fit-hypothesis or not.}
	\label{fig:concat_googleNet}
\end{figure}

\begin{figure}[t]
	\centering
	\scriptsize
	\includegraphics[width=\concatsize\linewidth]{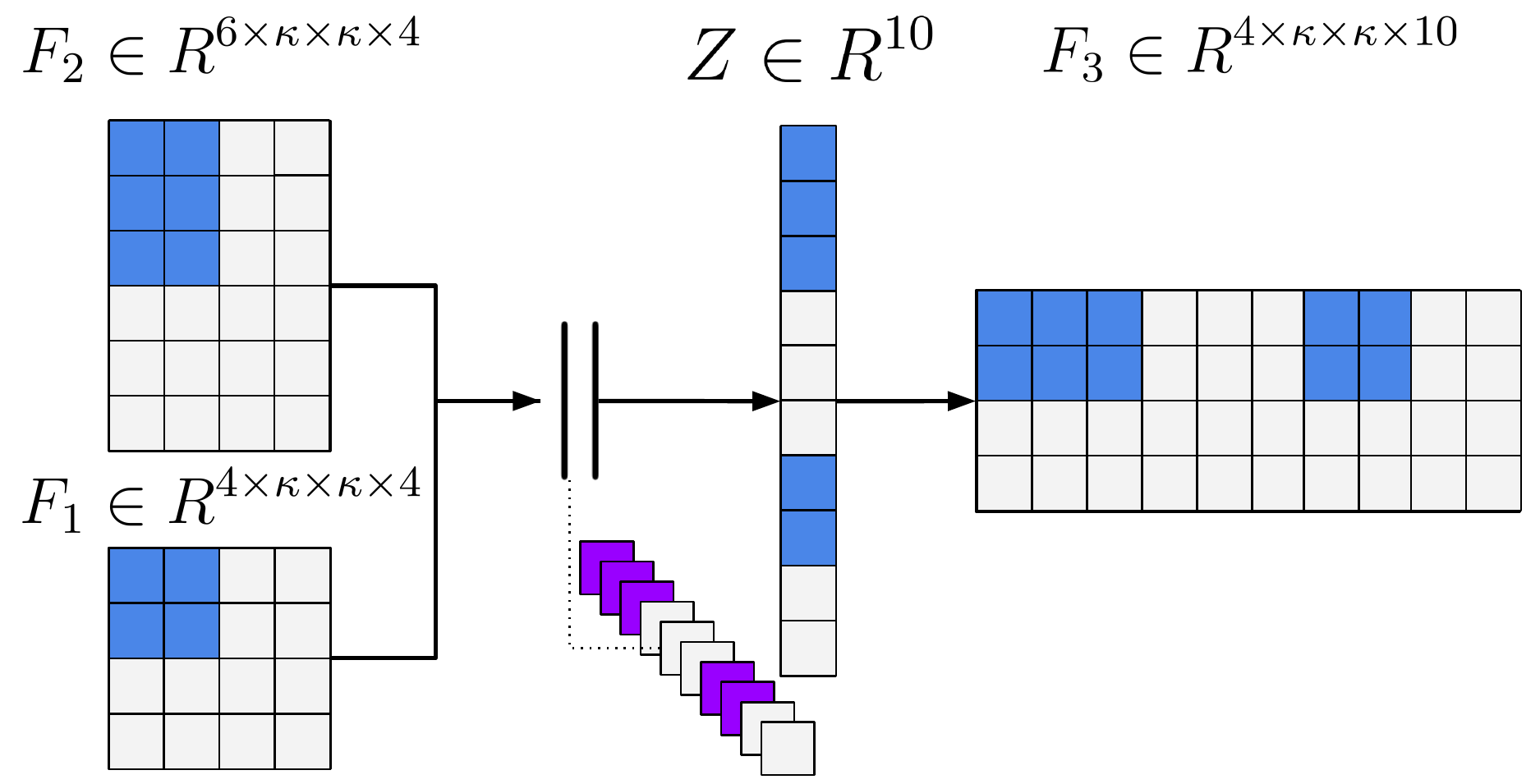}
	\caption{A Split-Net illustration on a toy feature concatenation operation. In this example, feature concatenation is followed by a batch norm layer then the convolutional layer $F_3$. This order of operations is employed in DenseNet.}
	\label{fig:concat_densenet}
\end{figure}

The kernel-level convolutional-aware splitting (KELS) technique supports both vanilla and residual networks. However, KELS requires a simple modification to support the concatenation operations (concat-op) in GoogLeNet and DenseNet. Figures~\ref{fig:concat_googleNet} and~\ref{fig:concat_densenet} illustrate how to handle concatenation in these networks. The main difference between \autoref{fig:concat_googleNet} and~\ref{fig:concat_densenet} is whether the concat-op is followed by a convolution or a batch-norm. To handle both variants,  we keep references to the preceding convolutional filters (\eg $F_1$ and $F_2$ in~\autoref{fig:concat_googleNet}). Using these references, we outline the fit-hypothesis in the convolutional and batch-norm layers. In this way, we split the network properly and make sure the fit-hypothesis is a slim independent network.


\keheading{Appendix  Intuition \#1: Dropout}
In the paper, we have illustrated how Split-Nets resemble dropout,~\ie both encourage neurons (subnetwork) to learn an independent representation. However, Split-Nets  target a specific set of neurons (subnetwork). For instance, if a toy network layer has 10 neurons, dropout promotes an independent representation to all 10 neurons. In contrast, Split-Nets  promote an independent representation to the neurons inside the fit-hypothesis $H^\triangle$ only. Thus, the split-mask $M$ provides a finer level of control. 




After highlighting the resemblance between KE and dropout, we want to emphasize that extending dropout for CNNs (channel-dropout) \textit{seems} trivial, but it is not.  Channel-dropout has been challenging because features in deep layers have great specificity~\cite{zeiler2014visualizing,yosinski2014transferable}. For an input image, a small fraction of channels is activated~\cite{zhang2016picking}. Thus, it is important to treat channels \textit{unequally},~\ie uniform random dropping is deficient. Consequently, Hou and Wang~\cite{hou2019weighted} have proposed Weighted Channel Dropout (WCD). This approach adds three extra modules to a deep network: Global Average Pooling, Weighted Random Selection, and Random Number Generator. These three modules are added to multiple convolutional layers.

Similar to KE, WCD~\cite{hou2019weighted} is applied during training. However, WCD does not reduce the inference cost. In addition, Hou and Wang~\cite{hou2019weighted} apply WCD to certain --not all -- convolutional layers (\eg \textit{res5a} and \textit{res5c} in ResNet-101). Thus, WCD requires tuning per architecture.\\


\keheading{Appendix Intuition \#2: Residual Network}
It is challenging to train a deep network on a small dataset. This challenge stems from the large number of parameters in a deep network. While all parameters are required for a large dataset, they become redundant and enable overfitting  on a small dataset. To mitigate overfitting, weight regularizers (\eg weight-decay) have been proposed. These regularizers  reduce the network's complexity by suppressing the weights' magnitudes,~\ie promote a \textit{zero-mapping}. 

A Res-Net splits a network into two branches: an identity shortcut and residual subnetwork. This network-splitting enables a \textit{zero-mapping} in residual links since a default identity mapping already exists. From this perspective, Res-Nets resemble weight-decay in terms of favoring a simpler subnetwork (\eg $R(x)=0; \forall x$). Yet, one difference is that a Res-Net can suppress the residual subnetworks while keeping the network's depth intact.





Similar to Res-Nets, a Split-Net splits a network into two branches: the fit-hypothesis $H^\triangle$ and the reset-hypothesis $H^\triangledown$. Split-Nets promote a zero mapping inside $H^\triangledown$ because, after the first generation, $H^\triangle$ is always closer to convergence. A zero-mapping inside $H^\triangledown$ reduces the number of active parameters, which in turn mitigates overfitting and reduces the burden for data collection. If all weights inside $H^\triangledown$ converge to zero, the network's depth remains intact,  thanks to the fit-hypothesis $H^\triangle$.





\section{Appendix:  Experiments}

\begin{table}[t]
	\scriptsize
	\centering
	\caption{The dimensions of the ResNet18 $N$ versus its fit-hypothesis $H^\triangle$ with split-rate $s_r=0.5$. The last table-section  compares $N$ and $H^\triangle$ through the number of operations and parameters (millions).
		The fit-hypothesis $H^\triangle$ is a slim independent network with 102 logits (Flower-102).}
	\begin{tabular}{@{}lll@{}}
		\toprule
		Layers                & ResNet18   $N$                       & Fit-hypothesis $H^\triangle$          \\
		
		\midrule
		conv1                             & 64 $\times$ 7 $\times$ 7 $\times$ 3    & 32 $\times$ 7 $\times$ 7 $\times$ 3    \\
		bn1                               & 64                                     & 32                                     \\
		layer1.0.conv1                    & 64 $\times$ 3 $\times$ 3 $\times$ 64   & 32 $\times$ 3 $\times$ 3 $\times$ 32   \\
		layer1.0.bn1                      & 64                                     & 32                                     \\
		layer1.0.conv2                    & 64 $\times$ 3 $\times$ 3 $\times$ 64   & 32 $\times$ 3 $\times$ 3 $\times$ 32   \\
		layer1.0.bn2                      & 64                                     & 32                                     \\
		layer1.1.conv1                    & 64 $\times$ 3 $\times$ 3 $\times$ 64   & 32 $\times$ 3 $\times$ 3 $\times$ 32   \\
		layer1.1.bn1                      & 64                                     & 32                                     \\
		layer1.1.conv2                    & 64 $\times$ 3 $\times$ 3 $\times$ 64   & 32 $\times$ 3 $\times$ 3 $\times$ 32   \\
		layer1.1.bn2                      & 64                                     & 32                                     \\
		layer2.0.conv1                    & 128 $\times$ 3 $\times$ 3 $\times$ 64  & 64 $\times$ 3 $\times$ 3 $\times$ 32   \\
		layer2.0.bn1                      & 128                                    & 64                                     \\
		layer2.0.conv2                    & 128 $\times$ 3 $\times$ 3 $\times$ 128 & 64 $\times$ 3 $\times$ 3 $\times$ 64   \\
		layer2.0.bn2                      & 128                                    & 64                                     \\
		layer2.0.downsample.0             & 128 $\times$ 1 $\times$ 1 $\times$ 64  & 64 $\times$ 1 $\times$ 1 $\times$ 32   \\
		layer2.0.downsample.1             & 128                                    & 64                                     \\
		layer2.1.conv1                    & 128 $\times$ 3 $\times$ 3 $\times$ 128 & 64 $\times$ 3 $\times$ 3 $\times$ 64   \\
		layer2.1.bn1                      & 128                                    & 64                                     \\
		layer2.1.conv2                    & 128 $\times$ 3 $\times$ 3 $\times$ 128 & 64 $\times$ 3 $\times$ 3 $\times$ 64   \\
		layer2.1.bn2                      & 128                                    & 64                                     \\
		layer3.0.conv1                    & 256 $\times$ 3 $\times$ 3 $\times$ 128 & 128 $\times$ 3 $\times$ 3 $\times$ 64  \\
		layer3.0.bn1                      & 256                                    & 128                                    \\
		layer3.0.conv2                    & 256 $\times$ 3 $\times$ 3 $\times$ 256 & 128 $\times$ 3 $\times$ 3 $\times$ 128 \\
		layer3.0.bn2                      & 256                                    & 128                                    \\
		layer3.0.downsample.0             & 256 $\times$ 1 $\times$ 1 $\times$ 128 & 128 $\times$ 1 $\times$ 1 $\times$ 64  \\
		layer3.0.downsample.1             & 256                                    & 128                                    \\
		layer3.1.conv1                    & 256 $\times$ 3 $\times$ 3 $\times$ 256 & 128 $\times$ 3 $\times$ 3 $\times$ 128 \\
		layer3.1.bn1                      & 256                                    & 128                                    \\
		layer3.1.conv2                    & 256 $\times$ 3 $\times$ 3 $\times$ 256 & 128 $\times$ 3 $\times$ 3 $\times$ 128 \\
		layer3.1.bn2                      & 256                                    & 128                                    \\
		layer4.0.conv1                    & 512 $\times$ 3 $\times$ 3 $\times$ 256 & 256 $\times$ 3 $\times$ 3 $\times$ 128 \\
		layer4.0.bn1                      & 512                                    & 256                                    \\
		layer4.0.conv2                    & 512 $\times$ 3 $\times$ 3 $\times$ 512 & 256 $\times$ 3 $\times$ 3 $\times$ 256 \\
		layer4.0.bn2                      & 512                                    & 256                                    \\
		layer4.0.downsample.0             & 512 $\times$ 1 $\times$ 1 $\times$ 256 & 256 $\times$ 1 $\times$ 1 $\times$ 128 \\
		layer4.0.downsample.1             & 512                                    & 256                                    \\
		layer4.1.conv1                    & 512 $\times$ 3 $\times$ 3 $\times$ 512 & 256 $\times$ 3 $\times$ 3 $\times$ 256 \\
		layer4.1.bn1                      & 512                                    & 256                                    \\
		layer4.1.conv2                    & 512 $\times$ 3 $\times$ 3 $\times$ 512 & 256 $\times$ 3 $\times$ 3 $\times$ 256 \\
		layer4.1.bn2                      & 512                                    & 256                                    \\
		fc                                & 102 $\times$ 512                       & 102 $\times$ 256                       \\
		\midrule
		\#Ops (G-Ops) & 3.63  &  0.96 \\
		\#Parameters & 22.44 &  5.64 \\
		\bottomrule
	\end{tabular}
	\label{tbl:split_resnet}
\end{table}

\subsection{Knowledge Evolution on Classification}

We have used public implementations for our baselines: RePr\footnote{\href{https://github.com/siahuat0727/RePr}{https://github.com/siahuat0727/RePr}}, BANs\footnote{\href{https://github.com/nocotan/born_again_neuralnet}{https://github.com/nocotan/born\_again\_neuralnet}}, AdaCos\footnote{\href{https://github.com/4uiiurz1/pytorch-adacos}{https://github.com/4uiiurz1/pytorch-adacos}}, and CS-KD\footnote{\href{https://github.com/alinlab/cs-kd}{https://github.com/alinlab/cs-kd}}. We leverage a public implementation\footnote{\href{https://github.com/mitchellnw/micro-net-dnw/blob/master/image_classification/model_profiling.py}{https://github.com/mitchellnw/micro-net-dnw/blob/master/image\_classification/model\_profiling.py}} to profile the fit-hypothesis computational cost. 

 In the paper manuscript,~\autoref{fig:ke_overview} illustrates the KELS technique on a toy Res-Net.~\autoref{tbl:split_resnet} uses  the ResNet18 architecture and a split-rate $s_r=0.5$ to present (1) the dimensions of both the dense network $N$ and the slim fit-hypothesis $H^\triangle$; (2) the computational cost of both $N$ and $H^\triangle$. The paper manuscript evaluates KE on DenseNet169 using the WELS technique and a split-rate $s_r=0.7$. Tables~\ref{tbl:densenet169_kels_08} and ~\ref{tbl:densenet169_wels_08} present quantitative classification evaluations on DenseNet169 using KELS and WELS, respectively. Both WELS and KELS evaluations use $s_r=0.8$.\\

In the paper manuscript, all experiments employ randomly initialized networks. Yet, pretrained networks achieve better performance on relatively small datasets.~\autoref{tbl:imagenet_performance} highlights the performance gap between randomly initialized (CS-KD+KE) and ImageNet initialized  (CE+ImageNet) networks. The CE+ImageNet baseline provides an upper bound. The CS-KD+KE baseline use KELS and $s_r=0.8$ with ResNet18, and WELS and $s_r=0.7$ with DenseNet169,~\ie last rows in Tables~\ref{tbl:resnet18} and~\ref{tbl:densenet169}. KE closes the performance gap between randomly initialized and ImageNet initialized networks significantly.\\

\begin{table}[t]
	\scriptsize
	\centering
	\caption{Quantitative evaluation using DenseNet169 with KELS and $s_r=0.8$,~\ie$\approx$ 36\% sparsity.}
	\begin{tabular}{@{}l c c c c c@{}}
		\toprule
		Method & Flower & CUB & Aircraft & MIT& Dog\\
		\midrule
		CE ($N_1$) & 45.76	& 55.49	& 51.96 &	57.37	& 65.09\\
		CE + KE-$N_3$ \textbf{(ours)}&  50.50 & 57.73	& 56.34	& 60.64	& 66.08\\
		CE + KE-$N_{10}$ \textbf{(ours)}&  \bf58.78	& \bf58.96	& \bf61.70	& \bf61.76	& \bf67.30 \\
		\midrule
		Smth ($N_{1}$)& 45.85 &	59.01	& 58.45	& 57.07	& 66.31\\
		Smth + KE-$N_{3}$ \textbf{(ours)}& 53.69	& \bf62.38	& 63.18	& \bf59.52	& 68.00 \\
		Smth + KE-$N_{10}$ \textbf{(ours)}& \bf65.88	& 60.57	& \bf65.60	& 59.15	& \bf68.66 \\
		\midrule
		CS-KD ($N_{1}$)& 49.32	& 66.71 &	57.62	&56.77 &	68.82\\
		CS-KD + KE-$N_{3}$ \textbf{(ours)}& 59.67	& \bf69.63	& 59.43	&57.14&	70.66 \\
		CS-KD + KE-$N_{10}$ \textbf{(ours)}& \bf66.34	& 69.35	& \bf59.76	& \bf 57.37&	\bf70.59 \\
		\bottomrule
	\end{tabular}
	\label{tbl:densenet169_kels_08}
\end{table}

\begin{table}[t]
	\scriptsize
	\centering
	\caption{Quantitative evaluation using DenseNet169 with WELS and split-rate $s_r=0.8$,~\ie 20\% sparsity.}
	\begin{tabular}{@{}l c c c c c@{}}
		\toprule
		Method & Flower  & CUB & Aircraft & MIT & Dog\\
		\midrule
		CE ($N_1$) & 44.88	& 56.32	& 51.61	& 55.13	& 66.15\\
		CE + KE-$N_3$ \textbf{(ours)}&  50.23	& \bf59.81 &	56.25	& \bf60.27 &	66.44\\
		CE + KE-$N_{10}$ \textbf{(ours)}&  \bf58.03 &	59.38	& \bf60.80	&59.45	& \bf67.25 \\
		\midrule
		Smth ($N_{1}$) 											& 45.92	& 58.70	&56.73	& 58.26	& 66.48\\
		Smth + KE-$N_{3}$ \textbf{(ours)}& 54.84 & 	\bf62.41	& 62.68	& 60.49	& 67.98 \\
		Smth + KE-$N_{10}$ \textbf{(ours)}& \bf64.69	& 60.36 &	\bf65.62	& \bf62.13	& \bf68.26 \\
		\midrule
		CS-KD ($N_{1}$) 										& 46.75	& 66.66	& 58.87	& 56.85	& 69.22 \\
		CS-KD + KE-$N_{3}$ \textbf{(ours)}& 58.27	& 69.67	& 60.98	& \bf57.51	& 70.94\\
		CS-KD + KE-$N_{10}$ \textbf{(ours)}& \bf64.18	& \bf71.37	& \bf61.37 &	57.22	& \bf71.33 \\
		\bottomrule
	\end{tabular}
	\label{tbl:densenet169_wels_08}
\end{table}

\begin{table}[t]
	\scriptsize
	\centering
	\caption{Comparative evaluation between pretrained (CE + ImageNet) and randomly initialized (CS-KD + KE) networks. The performance of CE + ImageNet provides an upper-bound for KE.}
	\begin{tabular}{@{}l c c c c c@{}}
		\toprule
		Method & Flower  & CUB & Aircraft & MIT & Dog\\
		\midrule
		& \multicolumn{5}{c}{ResNet18}\\ 						\cmidrule{2-6}
		CE + ImageNet & \bf 88.83	& \bf 74.46	& \bf61.01	& \bf72.84 & \bf 74.29\\
		CS-KD + KE-$N_{10}$	& 69.88	& 73.39	& 59.08	& 57.96	& 70.81\\ 
		\midrule
		& \multicolumn{5}{c}{DenseNet169}\\ 
		\cmidrule{2-6}
		CE  + ImageNet& \bf 93.46	& \bf 80.73	& \bf69.85	& \bf77.90 & \bf 79.92\\
		CS-KD + KE-$N_{10}$	& 65.27	& 70.36	& 61.22	& 57.44	& 70.72\\ 
		
		\bottomrule
	\end{tabular}
	\label{tbl:imagenet_performance}
\end{table}




\keheading{KE vs RePr} 
In the paper manuscript, we highlight two differences between KE and RePr. Yet, there are other worth noting differences.  (\rom{1})  RePr delivers a dense network only. (\rom{2}) RePr's re-initialization step (QR decomposition) is computationally expensive. (\rom{3}) During training, RePr prunes a different set of filters at different stages. If the pruned filters are regarded as a reset-hypothesis, then RePr changes the reset-hypothesis at different training stages. In contrast, KE outlines both fit and reset hypotheses using a single split-mask. This mask remains the same across all generations.\\

\keheading{KE vs DSD}
DSD is a prominent training approach. Han~\etal~\cite{han2016dsd} evaluated DSD using various tasks: image classification, caption generation, and speech recognition. Surprisingly, the DSD's intuition is never discussed in its paper~\cite{han2016dsd}. 


\colorlet{soulgray}{gray!40}
\sethlcolor{soulgray}  
We claim that DSD is a special case of KE. To support this claim, we first summarize the DSD training approach in ~\autoref{algo:dsd}. In this algorithm, we focus on two steps: Step \#8 and Step \#11. In \hl{Step \#8}, DSD outlines the less important weights to  be pruned using the binary variable $Mask$. This step is similar to our network-splitting step that outlines the fit and reset hypotheses through WELS. However, WELS splits a network $N$ randomly while DSD splits $N$ using a weight magnitude threshold.

\setcounter{algocf}{1}
\begin{algorithm}[t]
	\SetAlgoLined
	\KwResult{$W^{(t)}$}
	\BlankLine
	$W^{(0)} \sim  N(0,\Sigma )$\tcp*{\scriptsize{Randomly initialize} $W^{(0)}$} 
	\While(\tcp*[h]{Dense Phase}){not converged}
	{
		$W^{(t)} = W^{(t-1)}-  \eta ^{(t)} \nabla f(W^{(t-1)};x^{t-1})$\;
		$t = t+1$\;
	}
	\tcp{Sparse Phase} 
	$S = sort(abs(W^{(t-1)}))$  \tcp*{ \scriptsize{descendingly}} 
	$\lambda = S[k]$\;
	\hl{$\text{Mask} = \mathds{1}( abs(W^{(t-1)}) > \lambda)$}\;
	\While{not converged}
	{
		$W^{(t)} = W^{(t-1)}-  \eta ^{(t)} \nabla f(W^{(t-1)};x^{t-1})$\;
		\hl{$W^{(t)} =  W^{(t)} \text{Mask}$}\;
		$t = t+1$\;
	}
	\While(\tcp*[h]{Dense Phase}){not converged}
	{
		$W^{(t)} = W^{(t-1)}-  \eta ^{(t)} \nabla f(W^{(t-1)};x^{t-1})$\;
		$t = t+1$\;
	}
	\caption{Workflow of DSD from~\cite{han2016dsd}. The $\lambda = S[k]$ denotes the k-th largest weight where $k=|W|*(1-sparsity)$, and $|W|$ is the number of weights inside a network.}
	\label{algo:dsd}
\end{algorithm}

\hl{Step \#11} re-initializes the pruned weights to zero. Again, this step is similar to our reset-hypothesis re-initialization step. However, there are two differences. (1) We re-initialize the reset-hypothesis randomly instead of zero-values. If the re-initialization step is regarded as a noise injection process, then DSD injects noise with a zero standard deviation. In contrast, KE injects  noise with a non-zero standard deviation. This difference is important because the DSD's noise (zero-values) is bad for KELS. KELS re-initializes entire filters in the reset-hypothesis,~\ie a zero filter is an inferior initialization. (2) KE injects noise efficiently,~\ie across generations only. In contrast, DSD executes Step \#11 for every training mini-batch. Concretely, if we train a network on a dataset of size $B$, the re-initialization cost is $O(g \times L)$ for KE, and $O(g \times e \times  \frac{B}{b} \times L)$ for DSD, where $g$ is the number of generations, $e$ is the number of epochs, $L$ is the number of layers, and $b$ is the mini-batch size. The vanilla DSD assumes $g=1$, but this is an inferior setting as we show next.



\begin{figure}[t]
	\centering
	\scriptsize
	\begin{tikzpicture}
		\begin{axis}[
			xmin=1,
			xmax=10,
			width=0.7\linewidth,
			y label style={at={(axis description cs:0.15,.5)}},
			legend pos=outer north east,
			legend style={
				legend columns=1},
			ylabel=Top-1]
			
			\addplot[color=blue] coordinates {
				(1, 64.76)(2, 69.30)(3, 71.15)(4, 72.48)(5, 72.46)(6, 72.70)(7, 73.17)(8, 72.74)(9, 73.17)(10, 72.93)
			};\addlegendentry{KE+KELS $s_r=0.8$}

			\addplot[color=blue,dashed] coordinates {
				(1, 65.00)(2, 69.46)(3, 71.79)(4, 72.05)(5, 72.29)(6, 72.76)(7, 73.38)(8, 73.96)(9, 73.10)(10, 73.50)
			};\addlegendentry{KE+WELS $s_r=0.7$}
			
			\addplot[color=red,mark=diamond*] coordinates {
				(1, 64.45)(2, 69.70)(3, 72.10)(4, 72.27)(5, 72.86)(6, 73.41)(7, 73.39)(8, 74.59)(9, 73.43)(10, 73.84)
			};\addlegendentry{KE+DSD\space \space \space $s_r=0.7$}

		\end{axis}
	\end{tikzpicture}
	
	\begin{tikzpicture}
		\begin{axis}[
			xlabel=Generation \# ($g$),
			xmin=1,
			xmax=10,
			width=0.7\linewidth,
			y label style={at={(axis description cs:0.15,.5)}},
			legend pos=outer north east,
			legend style={
				legend columns=1},
			ylabel=Top-1
			]
			\addplot[color=blue,mark=*] coordinates {
				(1, 0.64)(2, 59.27)(3, 68.94)(4, 71.31)(5, 71.53)(6, 71.89)(7, 72.30)(8, 72.13)(9, 72.29)(10, 71.67)
			};\addlegendentry{KE+KELS $s_r=0.8$}
			
		\end{axis}
	\end{tikzpicture}
	\caption{Quantitative comparison between KE and KE+DSD. (Top) The classification performance of the dense network $N$. (Bottom) The performance of the slim fit-hypothesis $H^\triangle$. Through KELS, $H^\triangle$ achieves 71.67\% top-1 accuracy at $g=10$.}
	\label{fig:ks_dsd}
\end{figure}
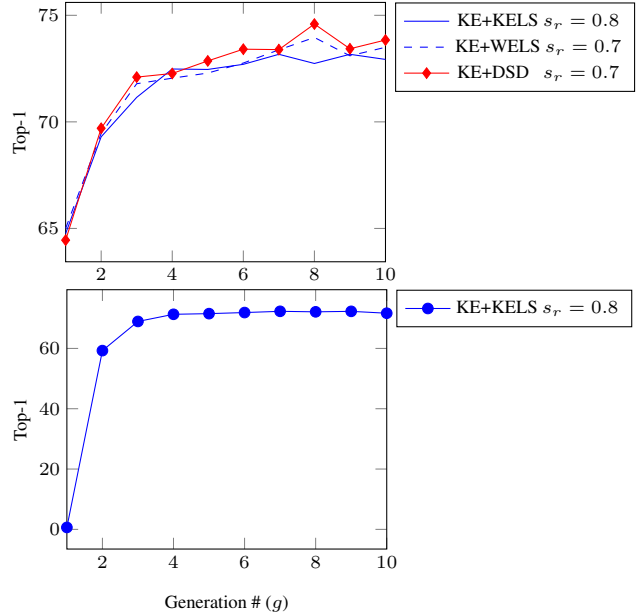

To highlight the similarity between KE and DSD quantitatively, we modify the vanilla DSD training approach. We keep the masking and re-initialization steps (Steps \#8 and \#11), but change the phases into generations. The dense and sparse phases become the old and \textit{even} generations, respectively. This modification means we do not \textit{resume} the learning rate $lr$ scheduler between phases, but \textit{re-start} the $lr$ scheduler across generations. Basically, we get rid of (1) the hard three-phases constraint, (2) the loss convergence criterion, and (3) the learning rate resumption across phases. We refer to this DSD variant as KE+DSD. Similar to KE, KE+DSD trains every generation for $e=200$ epochs.

\autoref{fig:ks_dsd} compares KE with our proposed KE+DSD. We train GoogLeNet for $g=10$ generations on CUB-200. We evaluate KE using both KELS and WELS. We use a split-rate $s_r=0.8$ with KELS and $s_r=0.7$ with WELS. For KE+DSD, we prune each layer  to the default 30\% sparsity.  KE+DSD achieves comparable performance to the KE. Yet, we want to highlight one  subtle difference between KE and KE+DSD. During training, KE allows all weights to change. However, KE+DSD freezes 30\% of the weights to zero at the even generations -- the original sparse phases -- through \hl{Step \#11}. This form of strict regularization gives KE+DSD a marginal edge during even generations -- the $8^{\text{th}}$ and the $10^{\text{th}}$ generations in \autoref{fig:ks_dsd}.


To conclude, DSD is a special case of KE. However, one clear difference between DSD~\cite{han2016dsd} and our paper is KELS. Through KELS, we learn both slim and dense networks simultaneously. Having said that, the main contribution of our paper is how we present a deep network as a set of hypotheses. We introduce the idea of a fit-hypothesis to encapsulate a network's knowledge. Then, we show how to evolve this knowledge to boost performance on relatively small datasets.




\subsection{Knowledge Evolution on Metric Learning}

\newcommand{\retchartsize}{0.55}
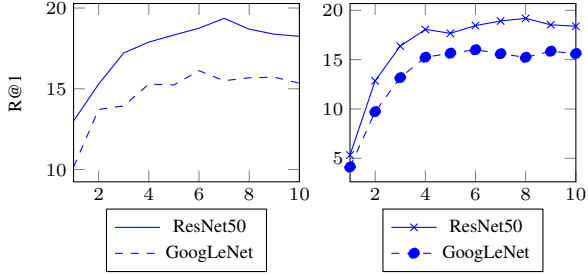
\begin{figure}[t]
	\centering
	\scriptsize
	\begin{tikzpicture}
		\begin{axis}[
			xmin=1,
			xmax=10,
			width=\retchartsize\linewidth,
			y label style={at={(axis description cs:0.15,.5)}},
			legend style={
				at={(0.5,-0.15)},
				anchor=north,
				legend columns=1},
			ylabel=R@1]
			
			\addplot[color=blue] coordinates {
				(1, 13.01)(2, 15.29)(3, 17.22)(4, 17.89)(5, 18.33)(6, 18.75)(7, 19.36)(8, 18.69)(9, 18.38)(10, 18.25)
			};\addlegendentry{ResNet50}
			
			\addplot[color=blue,dashed] coordinates {
				(1, 10.16)(2, 13.72)(3, 13.94)(4, 15.29)(5, 15.24)(6, 16.12)(7, 15.50)(8, 15.68)(9, 15.72)(10, 15.34)
			};\addlegendentry{GoogLeNet}
			
		\end{axis}
	\end{tikzpicture}\begin{tikzpicture}
		\begin{axis}[
			xmin=1,
			xmax=10,
			width=\retchartsize\linewidth,
			y label style={at={(axis description cs:0.15,.5)}},
			legend style={
				at={(0.5,-0.15)},
				anchor=north,
				legend columns=1},
			]
			\addplot[color=blue,mark=x] coordinates {
				(1, 5.33)(2, 12.85)(3, 16.37)(4, 18.06)(5, 17.67)(6, 18.45)(7, 18.92)(8, 19.18)(9, 18.53)(10, 18.38)
			};\addlegendentry{ResNet50}
			\addplot[color=blue,dashed,mark=*] coordinates {
				(1, 4.12)(2, 9.72)(3, 13.17)(4, 15.24)(5, 15.65)(6, 16.00)(7, 15.60)(8, 15.23)(9, 15.85)(10, 15.61)
			};\addlegendentry{GoogLeNet}
			
		\end{axis}
	\end{tikzpicture}
	\caption{Quantitative retrieval evaluation using CUB-200 on both GoogLeNet and ResNet50. Both networks are trained for 10 generations. (Left) Recall@1 of the dense network $N$. (Right) Recall@1 of the slim fit-hypothesis $H^\triangle$.
	}

	\label{fig:ret_cub}
\end{figure}

\ketopic{Evaluation Metrics} For metric learning evaluation, we leverage the \textbf{Recall@K} metric and \textbf{Normalized Mutual Info} (NMI) on the test split. The NMI score evaluates the quality of cluster alignments. $\text{NMI}=\frac { I(\Omega ,C) }{ \sqrt { H(\Omega )H(C) }  } ,$ where $\Omega =\{\omega_1,..,\omega_n\}$, is the ground-truth clustering, while $C=\{c_1,...c_n\}$ is a clustering assignment for the learned embedding. $I(\kebullet[0.5],\kebullet[0.5])$ and $H(\kebullet[0.5])$ denote mutual information and entropy, respectively. We use K-means to compute $C$. 

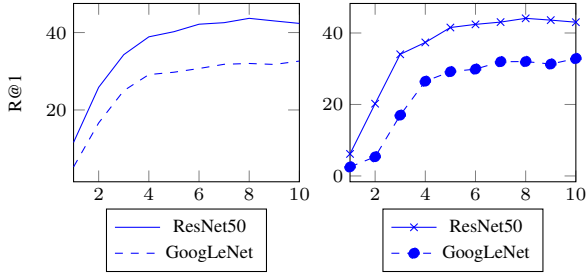
\begin{figure}[t]
	\centering
	\scriptsize
	\begin{tikzpicture}
		\begin{axis}[
			xmin=1,
			xmax=10,
			width=\retchartsize\linewidth,
			y label style={at={(axis description cs:0.15,.5)}},
			legend style={
				at={(0.5,-0.15)},
				anchor=north,
				legend columns=1},
			ylabel=R@1]
			
			\addplot[color=blue] coordinates {
				(1, 11.63)(2, 25.85)(3, 34.28)(4, 38.89)(5, 40.22)(6, 42.15)(7, 42.55)(8, 43.67)(9, 43.00)(10, 42.36)
			};\addlegendentry{ResNet50}
			
			\addplot[color=blue,dashed] coordinates {
				(1, 5.29)(2, 16.65)(3, 24.98)(4, 29.14)(5, 29.75)(6, 30.71)(7, 31.80)(8, 32.01)(9, 31.76)(10, 32.63)
			};\addlegendentry{GoogLeNet}
			
		\end{axis}
	\end{tikzpicture}\begin{tikzpicture}
		\begin{axis}[
			xmin=1,
			xmax=10,
			width=\retchartsize\linewidth,
			y label style={at={(axis description cs:0.15,.5)}},
			legend style={
				at={(0.5,-0.15)},
				anchor=north,
				legend columns=1},
			]
			\addplot[color=blue,mark=x] coordinates {
				(1, 6.17)(2, 20.22)(3, 34.04)(4, 37.36)(5, 41.53)(6, 42.41)(7, 43.05)(8, 44.10)(9, 43.60)(10, 43.02)
			};\addlegendentry{ResNet50}
			\addplot[color=blue,dashed,mark=*] coordinates {
				(1, 2.53)(2, 5.36)(3, 17.00)(4, 26.50)(5, 29.17)(6, 29.89)(7, 31.96)(8, 32.01)(9, 31.31)(10, 32.85)
			};\addlegendentry{GoogLeNet}
			
		\end{axis}
	\end{tikzpicture}
	\caption{Quantitative retrieval evaluation using CARS196 on both GoogLeNet and ResNet50. (Left) Recall@1 of the dense network $N$. (Right) Recall@1 of the slim fit-hypothesis $H^\triangle$.
	}
	
	\label{fig:ret_cars}
	\vspace{0.1in}
\end{figure}

\ketopic{Results} In the paper, we report the retrieval performance using the dense network $N$. However, KELS delivers a slim $H^\triangle$ as well. Figures~\ref{fig:ret_cub} and~\ref{fig:ret_cars} present quantitative retrieval evaluation on CUB-200 and CARS196, respectively. Both figures leverage the R@1 metric for quantitative evaluation. We report the performance of both the dense network $N$ and the slim fit-hypothesis $H^\triangle$. As the number of generations increases, the retrieval performance increases for both $N$ and $H^\triangle$.~\autoref{tbl:ke_kels_ret_eval} presents the fit-hypothesis $H^\triangle$ performance and inference cost. The fit-hypothesis $H^\triangle$ performance reaches the dense network $N$ performance after $g=10$ generations; yet, $H^\triangle$ achieves this performance at a significantly smaller inference cost.

\begin{table}[t]
	\centering
	\scriptsize
	\caption{Quantitative evaluation for KELS using the number of both operations (G-Ops) and parameters (millions). $\text{R1}_g$ denotes the recall@1 performance at the $g^{\text{th}}$ generation. $\blacktriangle_{\text{ops}}$ denotes the relative reduction in the number of operations. $\blacktriangle_{\text{r1}}$ denotes the absolute improvement margin on top of the dense baseline $N_1$.}
	\begin{tabular}{@{}lcrccccr@{}}
		\toprule
		&$s_r$&   $\text{R1}_1$ & $\text{R1}_{10}$  &$\blacktriangle_{\text{r1}}$ & \#Ops & $\blacktriangle_{\text{ops}}$ & \#Param \\ 				\cmidrule{2-8}
		& \multicolumn{7}{c}{CUB on GoogLeNet} \\
\cmidrule{2-8}
		$N_g $ 												  & \multirow{2}{*}{0.8}&     10.16   &   15.34  &5.1\%&  3.00  & -& 11.44 \\
		$H^\triangle_g $ & &  4.12     &   15.61   &5.4\%&  1.98  &34.0\%&  7.43 \\
		\midrule		
				& \multicolumn{7}{c}{CUB on ResNet50} \\
		\cmidrule{2-8}
		$N_g $ 												  & \multirow{2}{*}{0.8}&     13.01   &   18.25  &5.2\%&  8.19  & -& 47.48 \\
		$H^\triangle_g $ & &  5.33    &   18.38   &5.3\%&  5.32  &35.0\%&  30.55 \\

	\midrule
		& \multicolumn{7}{c}{CARS on GoogLeNet} \\
\cmidrule{2-8}
$N_g $ 												  & \multirow{2}{*}{0.8}&     5.29   &   32.63  &27.3\%&  3.00  & -& 11.44 \\
$H^\triangle_g $ & &  2.53     &   32.85   &27.5\%&  1.98  &34.0\%&  7.43 \\
\midrule		
& \multicolumn{7}{c}{CARS on ResNet50} \\
\cmidrule{2-8}
$N_g $ 												  & \multirow{2}{*}{0.8}&     11.63   &   42.36  &30.7\%&  8.19  & -& 47.48 \\
$H^\triangle_g $ & &  6.17    &   43.02   &31.3\%&  5.32  &35.0\%&  30.55 \\

		\bottomrule
	\end{tabular}
	\label{tbl:ke_kels_ret_eval}
	\vspace{-0.1in}
\end{table}



\section{Appendix: Ablation Study}
\begin{figure}[t]
	\centering
	\scriptsize

	\begin{tikzpicture}
		\begin{groupplot}[group style = {group size = 2 by 4, horizontal sep = 20pt}, 
			height=3.3cm,
			xmax=10,
			x label style={at={(axis description cs:0.5,-0.05)},anchor=north}
			]
			\nextgroupplot[title=\texttt{Conv-1}, 
			legend style = { legend columns = -1, legend to name = grouplegend,}]
			\addplot[	color=blue,	]
			coordinates {
				(1, 0.08493681252002716)(2, 0.11136557161808014)(3, 0.12285859882831573)(4, 0.12230779230594635)(5, 0.12317689508199692)(6, 0.12356199324131012)(7, 0.12555129826068878)(8, 0.1257077306509018)(9, 0.12555478513240814)(10, 0.126451313495636)
				
			}; 
			\addplot[color=red,dashed]
			coordinates {
				(1, 0.08755726367235184)(2, 0.04093746095895767)(3, 0.017116140574216843)(4, 0.012143289670348167)(5, 0.00989651121199131)(6, 0.008689271286129951)(7, 0.008561183698475361)(8, 0.008298011496663094)(9, 0.008697839453816414)(10, 0.008678192272782326)
				
			};
			\nextgroupplot[title=\texttt{Conv-2}, 
			legend style = { legend columns = -1, legend to name = grouplegend,}]
			\addplot[	color=blue,	]
			coordinates {
				(1, 0.014609038829803467)(2, 0.023442327976226807)(3, 0.026609638705849648)(4, 0.02683161571621895)(5, 0.027137508615851402)(6, 0.02731972560286522)(7, 0.02752663567662239)(8, 0.027643781155347824)(9, 0.02774996869266033)(10, 0.02799667976796627)
				
			}; 
			\addplot[color=red,dashed]
			coordinates {
				(1, 0.014834036119282246)(2, 0.007154650054872036)(3, 0.0033310363069176674)(4, 0.002240509493276477)(5, 0.0019245013827458024)(6, 0.001829108688980341)(7, 0.0018273280002176762)(8, 0.0018276867922395468)(9, 0.0018264935351908207)(10, 0.0018355001229792833)
			};

			\nextgroupplot[title=\texttt{Conv-3}, 
			legend style = { legend columns = -1, legend to name = grouplegend,}]
			\addplot[	color=blue,	]
			coordinates {
				(1, 0.009887644089758396)(2, 0.017364487051963806)(3, 0.019900618121027946)(4, 0.020257702097296715)(5, 0.020382963120937347)(6, 0.020495722070336342)(7, 0.020613806322216988)(8, 0.020733438432216644)(9, 0.02084914781153202)(10, 0.020969342440366745)
				
			}; 
			\addplot[color=red,dashed]
			coordinates {
				(1, 0.010663075372576714)(2, 0.004747464787214994)(3, 0.0019253201317042112)(4, 0.0013440673938021064)(5, 0.0012961566681042314)(6, 0.0012949412921443582)(7, 0.0012960736639797688)(8, 0.0012938895961269736)(9, 0.001295912661589682)(10, 0.0012937574647367)
				
			};
			\nextgroupplot[title=\texttt{Conv-4}, 
			legend style = { legend columns = -1, legend to name = grouplegend,}]
			\addplot[	color=blue,	]
			coordinates {
				(1, 0.007663301192224026)(2, 0.014404826797544956)(3, 0.01665215939283371)(4, 0.01706503890454769)(5, 0.017231592908501625)(6, 0.01737307198345661)(7, 0.01746041141450405)(8, 0.017528627067804337)(9, 0.017581263557076454)(10, 0.017632190138101578)
			}; 
			\addplot[color=red,dashed]
			coordinates {
				(1, 0.008415102027356625)(2, 0.0038613262586295605)(3, 0.0014980798587203026)(4, 0.0009610414854250848)(5, 0.0009160091285593808)(6, 0.0009169641998596489)(7, 0.0009150828118436038)(8, 0.0009145658113993704)(9, 0.0009150652913376689)(10, 0.0009157087770290673)
			};
			
			\nextgroupplot[title=\texttt{Conv-5}, 
			legend style = { legend columns = -1, legend to name = grouplegend,}]
			\addplot[	color=blue,	]
			coordinates {
				(1, 0.006976663134992123)(2, 0.013359373435378075)(3, 0.0157000832259655)(4, 0.016250496730208397)(5, 0.01653898134827614)(6, 0.016673484817147255)(7, 0.016793254762887955)(8, 0.01682431995868683)(9, 0.016910046339035034)(10, 0.016962140798568726)
			}; 
			\addplot[color=red,dashed]
			coordinates {
				(1, 0.007048910949379206)(2, 0.0037490795366466045)(3, 0.0015259113861247897)(4, 0.0009765581344254315)(5, 0.0009177231113426387)(6, 0.0009159065666608512)(7, 0.0009154105791822076)(8, 0.0009159633191302419)(9, 0.0009155141888186336)(10, 0.0009144711657427251)
			};
			\nextgroupplot[title=\texttt{Conv-6}, 
			legend style = { legend columns = -1, legend to name = grouplegend,}]
			\addplot[	color=blue,	]
			coordinates {
				(1, 0.005597176495939493)(2, 0.010805564001202583)(3, 0.012734858319163322)(4, 0.013187727890908718)(5, 0.013395411893725395)(6, 0.013496856205165386)(7, 0.01359627116471529)(8, 0.013610946014523506)(9, 0.013641825877130032)(10, 0.013646837323904037)
			}; 
			\addplot[color=red,dashed]
			coordinates {
				(1, 0.005656960885971785)(2, 0.0031590256839990616)(3, 0.0012368806637823582)(4, 0.0007260657730512321)(5, 0.0006521241157315671)(6, 0.0006477385759353638)(7, 0.0006467923521995544)(8, 0.0006468301871791482)(9, 0.0006467311177402735)(10, 0.0006469915970228612)
			};

			\nextgroupplot[title=\texttt{Conv-7}, 
			legend style = { legend columns = -1, legend to name = grouplegend,}]
			\addplot[	color=blue,	]
			coordinates {
				(1, 0.00649772584438324)(2, 0.012113806791603565)(3, 0.014121178537607193)(4, 0.014636305160820484)(5, 0.014803760685026646)(6, 0.014883006922900677)(7, 0.01492321863770485)(8, 0.014952638186514378)(9, 0.015008456073701382)(10, 0.0150249432772398)
			}; 
			\addplot[color=red,dashed]
			coordinates {
				(1, 0.006487805396318436)(2, 0.0040277643129229546)(3, 0.0019066743552684784)(4, 0.0010527627309784293)(5, 0.0007645592559129)(6, 0.0006922198808752)(7, 0.00066785654053092)(8, 0.0006527155637741089)(9, 0.0006478667492046952)(10, 0.0006474845577031374)
			};
			\nextgroupplot[title=\texttt{Conv-8}, 
			legend style = { legend columns = -1, legend to name = grouplegend,}]
			\addplot[	color=blue,	]
			coordinates {
				(1, 0.006869892124086618)(2, 0.011920956894755363)(3, 0.013233454897999763)(4, 0.01353289932012558)(5, 0.013599387370049953)(6, 0.013606086373329163)(7, 0.013626438565552235)(8, 0.013570631854236126)(9, 0.013544373214244843)(10, 0.013568565249443054)
			}; \addlegendentry{$\widehat{H}^\triangle$}	
			\addplot[color=red,dashed]
			coordinates {
				(1, 0.006895147264003754)(2, 0.003593746805563569)(3, 0.001979197608307004)(4, 0.0011880496749654412)(5, 0.0008306733798235655)(6, 0.0007111429003998637)(7, 0.0006744465790688992)(8, 0.0006547154043801129)(9, 0.0006482722819782794)(10, 0.0006478787399828434)
			};\addlegendentry{$\widehat{H}^\triangledown$}	
			
		\end{groupplot}
		\node[below] at ($(group c1r4.south) +(1.5,-0.25)$) {\pgfplotslegendfromname{grouplegend}}; 
	\end{tikzpicture}
	
	\caption{Quantitative evaluation using CUB-200 on VGG11\_bn. The x-axis denotes the number of generations. $\widehat{H}^\triangle$ and $\widehat{H}^\triangledown$ denote the mean absolute value inside $H^\triangle$ and $H^\triangledown$, respectively.}
	\label{fig:quan_cub_vgg11_full}
\end{figure}
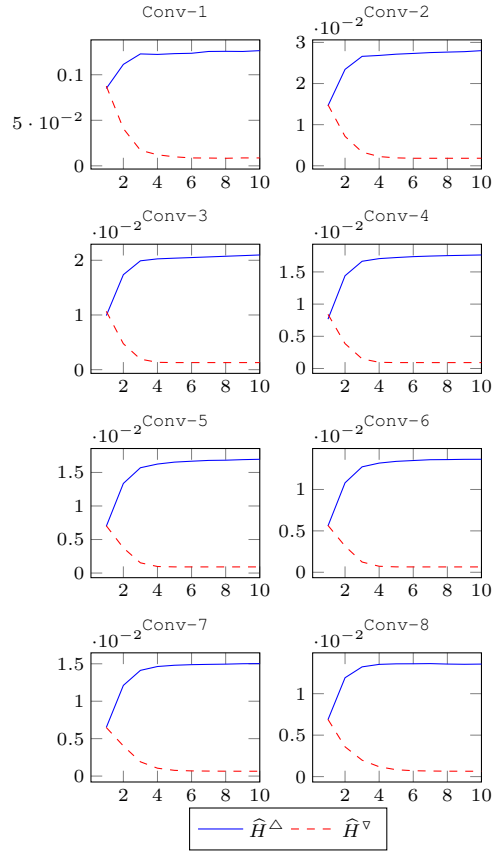

In the paper manuscript, we have utilized VGG11\_bn to monitor the development  of the fit and reset hypotheses across generations.~\autoref{fig:quan_cub_vgg11_full} shows the mean absolute values ($\widehat{H}^\triangle$ and $\widehat{H}^\triangledown$) inside the fit and reset hypotheses across all eight convolutional layers.

In this section, we present three ablation studies. We (1) evaluate the impact of changing the split-mask $M$ across generations, (2) discuss why the improvement-margins of KE differ among datasets, and (3) evaluate KE on a large dataset,~\ie ImageNet~\cite{deng2009imagenet}.

\keheading{(1) Changing the split-mask $M$ across generations}
In the paper manuscript, we split the network using a split-mark $M$. The \textit{same} mask is used to re-initialize every generation. However, we also highlighted the similarity between KE and dropout. Dropout does not drop the \textit{same} neurons during training. Thus, we investigate the impact of changing the split-mask $M$ across generations. This is possible with the WELS technique. In this experiment, We use CUB-200, ResNet18, label smoothing regularizer, the WELS technique, and four split-rates $s_r = \{0.2,0.3,0.5,0.8\}$. We train $N$ for 10 generations. After each generation, we re-initialize $M$ randomly,~\ie as if we initialize it for the first time. We refer to this WELS variant as WELS-Rand. 

\autoref{fig:hypothesis_reset} compares WELS against WELS-Rand. With small split-rates ($s_r = \{0.2,0.3\}$), WELS is significantly superior to WELS-Rand. However, as the split-rate increases ($s_r = \{0.5,0.8\}$), both WELS and WELS-Rand become comparable. This happens because different fit-hypotheses, in WELS+Rand, overlap partially. Given a split-rate $s_r$, a network-weight belongs to two consecutive fit-hypotheses with probability $s_r^2$. Accordingly, WELS-Rand with a small $s_r$ flushes the entire knowledge of a parent network. In contrast, WELS-Rand with a large split-rate retains the parent-network's knowledge  at least partially.

\newcommand{\resethypothesis}{0.8}
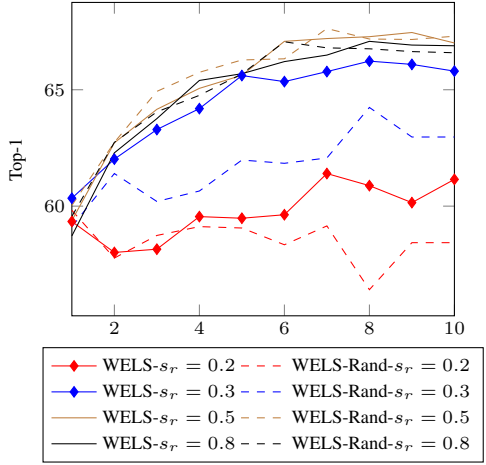
\begin{figure}[t]
	\centering
	\scriptsize
	\begin{tikzpicture}
		\begin{axis}[
			xmin=1,
			xmax=10,
			width=\resethypothesis\linewidth,
			y label style={at={(axis description cs:0.1,.5)}},
			ylabel=Top-1,
			legend style={
				at={(0.5,-0.1)},
				anchor=north,
				legend columns=2},
			]
			
			\addplot[color=red,solid,mark=diamond*] coordinates {
				(1, 59.34)(2, 58.01)(3, 58.15)(4, 59.55)(5, 59.48)(6, 59.63)(7, 61.40)(8, 60.88)(9, 60.15)(10, 61.15)
			};\addlegendentry{WELS-$s_r=0.2$}
			
			\addplot[color=red,dashed] coordinates {
				(1, 59.82)(2, 57.77)(3, 58.74)(4, 59.12)(5, 59.06)(6, 58.34)(7, 59.15)(8, 56.41)(9, 58.43)(10, 58.43)
			};\addlegendentry{WELS-Rand-$s_r=0.2$}		
			
			\addplot[color=blue,solid,mark=diamond*] coordinates {
				(1, 60.34)(2, 62.02)(3, 63.29)(4, 64.19)(5, 65.61)(6, 65.35)(7, 65.78)(8, 66.23)(9, 66.09)(10, 65.80)
			};\addlegendentry{WELS-$s_r=0.3$}
			
			\addplot[color=blue,dashed] coordinates {
				(1, 59.12)(2, 61.40)(3, 60.20)(4, 60.65)(5, 61.98)(6, 61.84)(7, 62.07)(8, 64.24)(9, 62.97)(10, 62.97)
			};\addlegendentry{WELS-Rand-$s_r=0.3$}		
			
			\addplot[color=brown,solid] coordinates {
				(1, 59.38)(2, 62.72)(3, 64.16)(4, 65.06)(5, 65.62)(6, 67.08)(7, 67.20)(8, 67.28)(9, 67.46)(10, 67.01)
			};\addlegendentry{WELS-$s_r=0.5$}
			
			\addplot[color=brown,dashed] coordinates {
				(1, 60.03)(2, 62.74)(3, 64.93)(4, 65.75)(5, 66.28)(6, 66.33)(7, 67.63)(8, 67.18)(9, 67.16)(10, 67.30)
			};\addlegendentry{WELS-Rand-$s_r=0.5$}
		
			\addplot[color=black,solid] coordinates {
(1, 58.72)(2, 62.29)(3, 63.76)(4, 65.40)(5, 65.69)(6, 66.21)(7, 66.49)(8, 67.08)(9, 66.92)(10, 66.89)
			};\addlegendentry{WELS-$s_r=0.8$}
		
				\addplot[color=black,dashed] coordinates {
			(1, 59.56)(2, 62.74)(3, 64.05)(4, 64.76)(5, 65.68)(6, 67.06)(7, 66.80)(8, 66.76)(9, 66.64)(10, 66.59)
		};\addlegendentry{WELS-Rand-$s_r=0.8$}



			
		\end{axis}
	\end{tikzpicture}

\caption{Comparative evaluation between WELS and WELS-Rand. WELS uses the same binary mask $M$ across all generations. In contrast, WELS-Rand randomly re-initialize $M$ after every generation. With a small split-rate, WELS-Rand flushes the parent-networks' knowledge.}
	\label{fig:hypothesis_reset}

\end{figure}

\keheading{(2) Why the improvement margins $\blacktriangle_{\text{acc}}$ of KE differ?}
In deep learning, we assume that more training data leads to better accuracy. However, the KE's improvement margins $\blacktriangle_{\text{acc}}$ contradict this assumption. For instance, \autoref{tbl:resnet18} shows that $\blacktriangle_{\text{acc}}$ on Flower-102  is bigger than $\blacktriangle_{\text{acc}}$ on CUB-200,~\ie $14.78$ vs $5.68$ after 10 generations with the CS-KD regularizer. \autoref{fig:intro_performance} also emphasizes this behavior; Flower-102 is a much smaller dataset compared to CUB-200, yet $\blacktriangle_{\text{acc}}$ is over 20\% for Flower-102 but less than 10\% for CUB-200. We posit that $\blacktriangle_{\text{acc}}$ depends not only on the dataset size, but also on the dataset simplicity. 



\begin{table}[t]
	\centering
	\scriptsize
	\caption{The KE's improvement margins $\blacktriangle_{\text{acc}}$ versus the FCAMD accuracies on each dataset. There is a strong positive Pearson correlation $(r=0.9529)$ between $\blacktriangle_{\text{acc}}$  and the datasets' simplicity (FCAMD's accuracies).}
	\begin{tabular}{@{}lcc@{}}
		\toprule
		Datasets & $\blacktriangle_{\text{acc}}$ & FCAMD Acc \\
		\midrule
		Flower & 14.78     & 63.06       \\
		CUB & 5.68	& 19.60  \\
		Aircraft & 0.93	& 15.80  \\
		MIT &           0.59	& 19.20       \\
		Stanford Dogs &          1.21	& 13.20       \\
		&& $r=0.952$ \\
		\bottomrule
	\end{tabular}
	\label{tbl:dataset_simplicity}
\end{table}

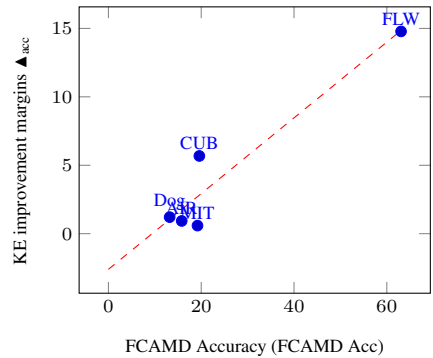
\begin{figure}[t]
	\scriptsize
	\centering
	\begin{tikzpicture}
		\begin{axis}[nodes near coords,
			width=0.75\linewidth,
			domain=0:63,
			xlabel={FCAMD Accuracy (FCAMD Acc)},
			ylabel={KE improvement margins $\blacktriangle_{\text{acc}}$},
			y label style={at={(axis description cs:0.1,.5)}},
			point meta=explicit symbolic]
			
			\addplot+[color=blue,only marks] coordinates {
				(63.066667,14.78)[FLW]
				(19.6,5.68)[CUB]
				(15.8,0.93)[AIR]
				(19.2,0.59)[MIT]
				(13.2,1.21)[Dog]
			};
			\addplot+[no marks,red,dashed] {0.2767636702 * x -2.605827812} ;
		\end{axis}%
	\end{tikzpicture}%
\caption{The average accuracy of the Flower (FLW), CUB, Aircraft (AIR), MIT, and Dog datasets inside the FCAMD dataset. The five datasets are equally represented inside FCAMD,~\ie 50 classes each and 10 images per class. The accuracy metric reflects the simplicity of each dataset. The x-axis denotes the accuracy of a dataset inside FCAMD and the y-axis denotes the KE improvement margins. There is a strong positive correlation between the datasets' simplicity and the KE improvement margins.}
	\label{fig:dataset_simplicity}
\end{figure}

To evaluate our postulate, we quantify the simplicity of our five datasets (\textbf{F}lower, \textbf{C}UB, \textbf{A}ircraft, \textbf{M}IT, and \textbf{D}og). We create a new dataset, dubbed FCAMD, using the five datasets. We randomly sample 50 classes from each dataset. For each class, we randomly sample 10 training and 10 testing images. Thus, FCAMD has 2500 training and 2500 testing images,~\ie 250 classes, 10 training images per class. We train a ResNet18 from scratch on FCAMD. To quantify the simplicity of each dataset, we measure the average accuracy of its 50 classes. Higher accuracy indicates  a simpler dataset. There is a strong positive Pearson correlation $(r=0.9529)$ between the datasets' simplicity (from FCAMD's accuracies) and the KE improvement margins $\blacktriangle_{\text{acc}}$ as shown in~\autoref{fig:dataset_simplicity} and~\autoref{tbl:dataset_simplicity}. To compute the Pearson correlation, we use the KE improvement margins $\blacktriangle_{\text{acc}}$ achieved after 10 generations on top of the CS-KD~\cite{yun2020regularizing} baseline,~\ie $\blacktriangle_{\text{acc}}$ from the last section of ~\autoref{tbl:resnet18}. Even if   we dismissed Flower-102 as an outlier, the correlation would become $r=0.494$ for the remaining four datasets (CUB, AIR, MIT, and Dog).

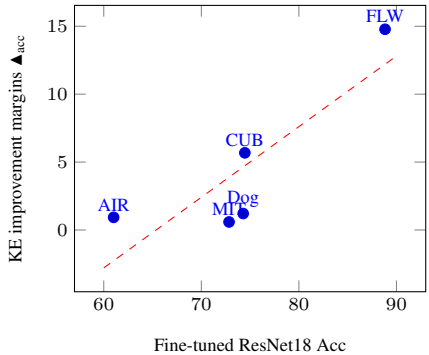
\begin{figure}[t]
	\scriptsize
	\centering	
	\begin{tikzpicture}
		\begin{axis}[nodes near coords,
			width=0.75\linewidth,
			domain=60:90,
			xlabel={Fine-tuned ResNet18 Acc},
			ylabel={KE improvement margins $\blacktriangle_{\text{acc}}$},
			y label style={at={(axis description cs:0.1,.5)}},
			point meta=explicit symbolic]
			
			\addplot+[color=blue,only marks] coordinates {
				(88.83,14.78)[FLW]
				(74.46,5.68)[CUB]
				(61.01,0.93)[AIR]
				(72.84,0.59)[MIT]
				(74.29,1.21)[Dog]
			};
			\addplot+[no marks,red,dashed] {0.520012756 * x -33.9916676} ;
		\end{axis}%
	\end{tikzpicture}%
	\caption{The  accuracy of the Flower (FLW), CUB, Aircraft (AIR), MIT, and Dog datasets on a \textit{fine-tuned} ResNet18. The accuracy metric reflects the simplicity of each dataset. The x-axis denotes the accuracy of a dataset on a  \textit{fine-tuned} ResNet18 and the y-axis denotes the KE improvement margins. There is a strong positive correlation between the datasets' simplicity and the KE improvement margins.}
	\label{fig:dataset_simplicity_imagenet}
\end{figure}

 Another way to quantify the simplicity of a dataset is through a pretrained network. A pretrained network contains the ImageNet's knowledge. This large knowledge mitigates the impact of both a small dataset size and a small number of samples per class. Thus, we fine-tune a pretrained ResNet18 on the five datasets as shown in~\autoref{tbl:imagenet_performance}. The accuracy of the fine-tuned ResNet18 reflects the simplicity of each dataset.  Higher accuracy indicates  a simpler dataset. Again, there is a strong positive Pearson correlation $(r=0.850)$ between $\blacktriangle_{\text{acc}}$ and the fine-tuned ResNet18 accuracies as shown in~\autoref{fig:dataset_simplicity_imagenet} and~\autoref{tbl:dataset_simplicity_imagenet}.

\begin{table}[t]
	\centering
	\scriptsize
	\caption{The KE's improvement margins $\blacktriangle_{\text{acc}}$ versus the accuracies of a \textit{fine-tuned} ResNet18. There is a strong positive Pearson correlation $(r=0.850)$ between  $\blacktriangle_{\text{acc}}$ and the datasets' simplicity (fine-tuned ResNet18 accuracies).}
	\begin{tabular}{@{}lcc@{}}
		\toprule
		Datasets & $\blacktriangle_{\text{acc}}$ & Fine-tuned ResNet18\\
		\midrule
		Flower & 14.78     & 88.83      \\
		CUB & 5.68	& 74.46  \\
		Aircraft & 0.93	& 61.01  \\
		MIT &           0.59	& 72.84       \\
		Stanford Dogs &          1.21	& 74.29       \\
		&& $r=0.850$ \\
		\bottomrule
	\end{tabular}
	\label{tbl:dataset_simplicity_imagenet}
\end{table}

The FCAMD and fine-tuned ResNet18 experiments present an interesting finding. It seems that the dataset size is no longer the dominant factor that controls  the performance of a randomly initialized network on relatively small datasets.

\keheading{(3) Evaluate KE on ImageNet}
Our paper tackles the following question: how to train a deep network on a relatively small dataset? Answering this question will have a significant impact on both academia and industry. However, it is important to understand how KE behaves on a large dataset,~\ie ImageNet. The goal of this experiment is \textit{not} to boost performance on ImageNet; Stock~\etal~\cite{stock2018convnets} and Beyer~\etal~\cite{beyer2020we} deliver strong arguments why boosting performance on ImageNet should no longer be an ultimate goal. While KE boosts performance on ImageNet, our goal is to monitor the performance of the fit-hypothesis. We want to answer the following question: can KE evolve knowledge inside the fit-hypothesis even when presented with a large dataset?

\ketopic{Technical Details} We train a ResNet18 for 5 generations using KELS and a split-rate $s_r=0.8$,~\ie $\approx 36\%$ sparsity. Our implementation for ImageNet follows the practice
in~\cite{he2016deep}. We use a batch size $b=128$, and a step learning rate scheduler with a starting $lr=0.1$. We train for $e=150$ epochs per generation. Other parameters (\eg momentum, optimizer) are the same as those reported in the paper (\autoref{sec:exp_cls}).\\



\ketopic{Results} \autoref{fig:cls_imagenet} presents a quantitative classification evaluation using ImageNet. KE boosts performance for both the dense network $N$ and the slim fit-hypothesis $H^\triangle$. In the paper manuscript, we evaluate KE using relatively small datasets and large architectures. In contrast, this experiment evaluates KE using a large dataset and a small architecture. Accordingly, these improvement margins on ImageNet are a lower-bound on the potential of KE. As the architecture gets bigger, these improvement margins will increase. Accordingly, we conclude that KE can evolve knowledge inside the fit-hypothesis.

\newcommand{\clschartsize}{0.55}
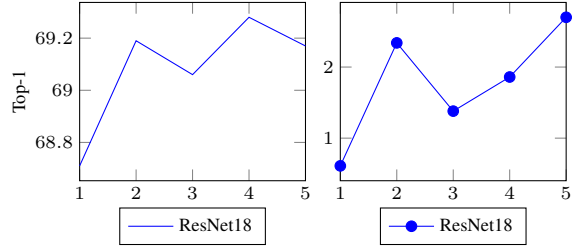
\begin{figure}[t]
	\centering
	\scriptsize
	\begin{tikzpicture}
		\begin{axis}[
			xmin=1,
			xmax=5,
			width=\clschartsize\linewidth,
			y label style={at={(axis description cs:0.15,.5)}},
			legend style={
				at={(0.5,-0.15)},
				anchor=north,
				legend columns=1},
			ylabel=Top-1]
			
			\addplot[color=blue] coordinates {
				(1, 68.71)(2, 69.19)(3, 69.06)(4, 69.28)(5, 69.17)
			};\addlegendentry{ResNet18}

		\end{axis}
	\end{tikzpicture}\begin{tikzpicture}
		\begin{axis}[
			xmin=1,
			xmax=5,
			width=\clschartsize\linewidth,
			y label style={at={(axis description cs:0.15,.5)}},
			legend style={
				at={(0.5,-0.15)},
				anchor=north,
				legend columns=1},
			]
			\addplot[color=blue,mark=*] coordinates {
				(1, 0.61)(2, 2.34)(3, 1.38)(4, 1.86)(5, 2.70)
			};\addlegendentry{ResNet18}
			
		\end{axis}
	\end{tikzpicture}
	\caption{Quantitative classification evaluation using ImageNet on ResNet18 for 5 generations. (Left) The accuracy performance (Top-1~$\uparrow$) of the dense network $N$. (Right) The performance of the slim fit-hypothesis $H^\triangle$.
	}

	\label{fig:cls_imagenet}
\end{figure}

\begin{table}[t]
	\centering
	\scriptsize
	\setlength\tabcolsep{5.50pt} 
	\caption{Quantitative classification evaluation using both ResNet34 and ResNet50. $N_g$ and $H^\triangle_g$  denote the performance of the dense network $N$ and the fit-hypothesis $H^\triangle$ at the $g^{\text{th}}$ generation. $\blacktriangle_{H}$ denotes the absolute improvement margin in the fit-hypothesis relative to the baseline $H^\triangle_1$}
	
	\begin{tabular}{@{}l   ccc l@{\hspace{1.0\tabcolsep}} ccc@{}}
		\toprule
		&  \multicolumn{3}{c}{ResNet34} && \multicolumn{3}{c}{ResNet50}\\
		\cmidrule{2-4} \cmidrule{6-8}
		g &  $N_g$ & $H^\triangle_g$ & $\blacktriangle_{H}$  &&  $N_g$ & $H^\triangle_g$ & $\blacktriangle_{H}$\\
		\midrule
		

		1        						  &   72.51		& 0.28	& -  &  &   74.54	&0.20&	-     \\
		2  \textbf{(ours)}&     \bf{72.86}	& 1.25	& \fpeval{1.25-0.28}     & &    74.78	& 3.44 &	\fpeval{3.44-0.20}    \\
		3  \textbf{(ours)}&     72.78	& 2.27	& \fpeval{2.27-0.28}    &&  75.01	& 6.71	& \fpeval{6.71-0.20}     \\
		4  \textbf{(ours)}&     \bf72.86	& 1.96	& \fpeval{1.96-0.28}    &&  75.15	& 4.63	& \fpeval{4.63-0.20}     \\
		5  \textbf{(ours)}&     \bf72.86	& \bf{4.49}	& \fpeval{4.49-0.28}    &&  \bf75.27	& \bf13.81	& \fpeval{13.81-0.20}     \\
		\bottomrule
	\end{tabular}
	
	\label{tbl:cls_imagenet_big_arch}
	\vspace{-0.02in}
\end{table}

We further evaluate KE on two larger architectures.~\autoref{tbl:cls_imagenet_big_arch} presents quantitative classification evaluation using ResNet34 and ResNet50. We use the same technical details from the ResNet18 experiment. KE boosts performance on the fit-hypothesis $H^\triangle$ consistently. This confirms our finding that KE evolves knowledge in the fit-hypothesis $H^\triangle$.

\end{document}